\documentclass{article}

\PassOptionsToPackage{numbers,compress}{natbib}
\usepackage[preprint]{neurips_2026}

\usepackage[utf8]{inputenc} 
\usepackage[T1]{fontenc}    
\usepackage{hyperref}       
\usepackage{url}            
\usepackage{booktabs}       
\usepackage{amsfonts}       
\usepackage{nicefrac}       
\usepackage{microtype}      
\usepackage{xcolor}         
\usepackage{multirow}
\usepackage{booktabs}
\usepackage{graphicx}
\usepackage{amsmath}
\usepackage{wrapfig}
\usepackage{subcaption}
\usepackage[dvipsnames]{xcolor}
\usepackage{enumitem}
\title{Overcoming Dynamics-Blindness: Training-Free Pace-and-Path Correction for VLA Models}

\author{%
  Yanyan Zhang\textsuperscript{1} \quad
  Chaoda Song\textsuperscript{1} \quad
  Vikash Singh\textsuperscript{1} \quad
  Xinpeng Li\textsuperscript{1} \quad
  Kai Ye\textsuperscript{1} \\
  {\bf Zhe Hu\textsuperscript{2} \quad
  Zhongzhu Pu\textsuperscript{3,4} \quad
  Yu Yin\textsuperscript{1} \quad
  Vipin Chaudhary\textsuperscript{1}} \\[0.3em]
  \textsuperscript{1}Case Western Reserve University \quad
  \textsuperscript{2}The Hong Kong Polytechnic University \quad \\
  \textsuperscript{3}Tsinghua University \quad
   \textsuperscript{4}InspireOmni AI\\[0.3em]
  \texttt{yxz3106@case.edu}
}

\begin{document}

\maketitle

\begin{abstract}
Vision-Language-Action (VLA) models achieve remarkable flexibility and generalization beyond classical control paradigms. However, most prevailing VLAs are trained under a single-frame observation paradigm, which leaves them structurally blind to temporal dynamics. Consequently, these models degrade severely in non-stationary scenarios, even when trained or finetuned on dynamic datasets. Existing approaches either require expensive retraining or suffer from latency bottlenecks and poor temporal consistency across action chunks. We propose Pace-and-Path Correction, a training-free, closed-form inference-time operator that wraps any chunked-action VLA. From a single quadratic cost, joint minimization yields a unified solution that decomposes orthogonally into two distinct channels. The pace channel compresses execution along the planned direction, while the path channel applies an orthogonal spatial offset, jointly absorbing the perceived dynamics within the chunk window. We evaluate our approach on a comprehensive diagnostic benchmark \textsc{MoveBench} designed to isolate motion as the sole controlled variable. Empirical results demonstrate that our framework consistently outperforms state-of-the-art training-free wrappers and dynamic-adaptive methods and improves success rates by up to 28.8\% and 25.9\% in absolute terms over foundational VLA models in dynamic-only and static-dynamic mixed environments, respectively.
\end{abstract}

\section{Introduction}

Robotic manipulation in real-world settings frequently involves environments whose state changes during policy execution, ranging from regular motions such as objects on a conveyor belt to unexpected events such as external perturbations~\citep{Zhang2025DynamicBC,Fang2026TowardsGR,Xie2026DynamicVLAAV}. Handling such dynamic conditions has therefore become a central requirement for general-purpose manipulation policies~\citep{Hu2023TowardGR,Ma2024ASO}. Among recent approaches, Vision-Language-Action (VLA) models map visual observations and language instructions directly to low-level control, and have emerged as a promising candidate for this setting~\citep{Team2024OctoAO,Brohan2023RT2VM,Kim2024OpenVLAAO}. 

However, most current VLAs adopt action chunking, where the model predicts a fixed-length sequence of future actions from a single visual frame at each inference call and the robot executes them open-loop before the next chunk is generated~\citep{pi0,act}. While this design improves stability and amortizes inference cost, it leaves the policy structurally blind to dynamics~\citep{Fang2026TowardsGR,Chi2023DiffusionPV}. Each chunk is generated from an initial static snapshot without object-motion supervision and executed without visual feedback, leaving any scene changes during execution unseen until the next chunk is generated~\citep{Brohan2023RT2VM,Nvidia2025GR00TNA}. As a result, even state-of-the-art VLAs that excel on static tasks degrade sharply once the task itself demands temporal awareness. Beyond methods, the evaluation landscape itself offers limited support for diagnosing motion robustness. Existing manipulation benchmarks rarely isolate motion as a primary axis, instead entangling it with perception, generalization, or scene difficulty, which makes the dynamics-blindness failure mode hard to characterize precisely~\citep{Fang2026TowardsGR,Xie2026DynamicVLAAV,Liu2023LIBEROBK,Mees2021CALVINAB}.

A growing body of recent work targets this gap, broadly falling into two strands. One injects motion or temporal cues into the input through historical optical flow~\citep{zhong2025flowvla,Fang2026TowardsGR,Fang2025RoboticVB}, visual prompting~\citep{Zheng2024TraceVLAVT}, memory banks~\citep{Shi2025MemoryVLAPM}, or motion predictors~\citep{Zhang2025DreamVLAAV,Cen2025WorldVLATA,Zhang20254DVLASV}, but these methods rely on expensive retraining and per-backbone architectural changes. Extraction latency and forecasting hallucinations make these methods unreliable at the timescale of dynamic interaction~\citep{zhong2025flowvla,Sendai2025LeaveNO,Jiang2026WoVRWM}. More fundamentally, a manipulator's visual stream is dominated by its own ego-motion, leaving genuine object motion as a small residual hard to capture~\citep{Zhi20253DFlowActionLC,Kambara2026LILACLO}. A second strand reduces inference latency through compact 
backbones~\citep{Xie2026DynamicVLAAV}, parallel 
decoding~\citep{Song2025PDVLAAV}, or compressed action 
tokenizers~\citep{Liu2025FASTerTE}, trading away the backbone capacity 
that gives larger VLAs their generalization while still leaving each 
newly issued chunk blind to motion within the previous one. Indiscriminate re-inference can also break the temporal smoothness across chunks and degrade long-horizon coherence~\citep{Black2025RealTimeEO}. Other methods include asynchronous inpainting~\citep{Black2025RealTimeEO}, 
rejection sampling~\citep{Liu2024BidirectionalDI}, temporal 
ensembling~\citep{act}, adaptive chunk sizing~\citep{Wen2026AdaptiveAC}, 
and learned correction heads~\citep{Sendai2025LeaveNO}, which improve 
reactivity indirectly through smoother seams or more frequent re-planning. However, the chunks themselves still treat the environment as static, and any learnable corrector still suffers from the dilemma between latency and capacity as well as the ego-motion problem~\citep{act,Huang2026TICVLAAT}. Without external dynamics information, identical initial observations with different target velocities make intra-chunk correction underdetermined.

\begin{figure}[t]
    \centering
    \includegraphics[width=\textwidth]{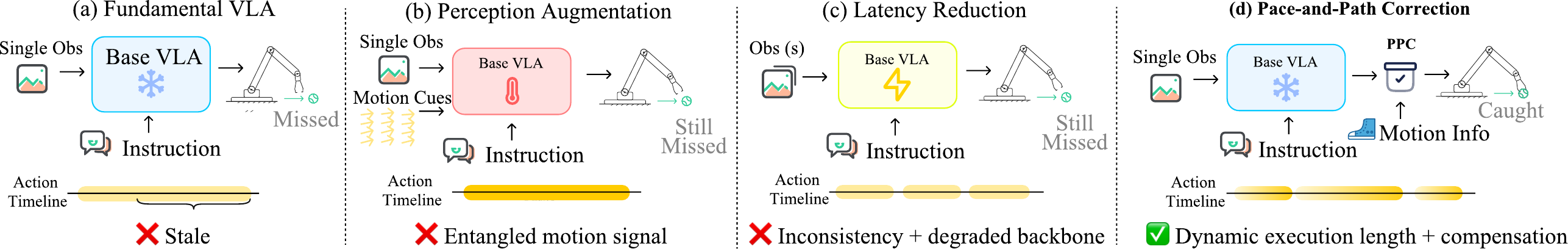}
    \caption{\textbf{Comparison of methods.} (a) Fundamental VLA suffers from single-frame input that leaves the latter half of each chunk stale under dynamic scenes. (b) Perception augmentation requires retraining, and the motion signal is progressively diluted through the VLA stack and ego-motion. (c) Latency reduction blindly accelerates inference, breaking chunk-to-chunk consistency and typically relying on a lightweight backbone. (d) Our framework adaptively compresses action magnitude and inference cadence, compensating spatially against the environment.}
    \label{fig:intro}
    \vspace{-1em}
\end{figure}

Therefore, we propose \textbf{Pace-and-Path Correction (PPC)}, a closed-form, training-free, inference-time wrapper. PPC reads an external dynamics signal in the form of velocity, which can be supplied by external tracking or depth-sensing pipelines. As illustrated in Fig.~\ref{fig:intro}, unlike prior remedies that augment the input, shrink the backbone, or smooth chunk boundaries, PPC directly addresses the chunk interior where dynamics blindness actually resides, through a principled, physics-grounded formulation. It solves a single quadratic cost balancing per-waypoint tracking against per-step offset effort in closed form, whose minimum decomposes orthogonally into two channels. \textbf{Pace} adaptively compresses the chunk in time to absorb the plan-parallel component of the disturbance, while \textbf{Path} adds per-step spatial offsets to absorb the plan-perpendicular component. A \textbf{Hierarchical 2-EMA Latch Stabilizer} further detects motion regimes and shortens the execution horizon for necessity under chronic instability. By decoupling perception from correction, PPC inherits the maturity of dedicated tracking pipelines, sidesteps the latency-capacity dilemma that constrains any learnable corrector, and avoids the ego-motion confound that handicaps in-backbone perception. The resulting wrapper is agnostic to the underlying backbone, requires negligible compute, and recovers the baseline VLA exactly under static environment, preserving the strong static-scene capability of modern foundational VLAs. To rigorously study PPC and the broader question of motion robustness, we further construct \textbf{\textsc{MoveBench}}, a controlled benchmark that isolates motion regime as the primary evaluation axis while holding tasks, objects, and scenes fixed. The key contributions of our work are summarized as follows:

\begin{itemize}[nosep,leftmargin=1.3em]
    \item We propose \textbf{Pace-and-Path Correction (PPC)}, a closed-form, training-free, inference-time wrapper for general VLAs that explicitly compensates for environment dynamics with no learnable parameters and no backbone modification or specification.

    \item We construct \textbf{\textsc{MoveBench}}, a benchmark dedicated to systematically isolating and evaluating VLA performance across diverse motion patterns and speeds.
   
    \item Extensive experiments demonstrate that PPC outperforms state-of-the-art training-free wrappers and dynamic-adaptive methods, and consistently enhances all motion families across various foundational VLAs, improving success rates by up to $28.8\%$ and $25.9\%$ in dynamic-only and mixed environments, respectively.
\end{itemize}

\section{Related Work}

\begin{figure*}[!t]
    \centering
    \includegraphics[width=\textwidth]{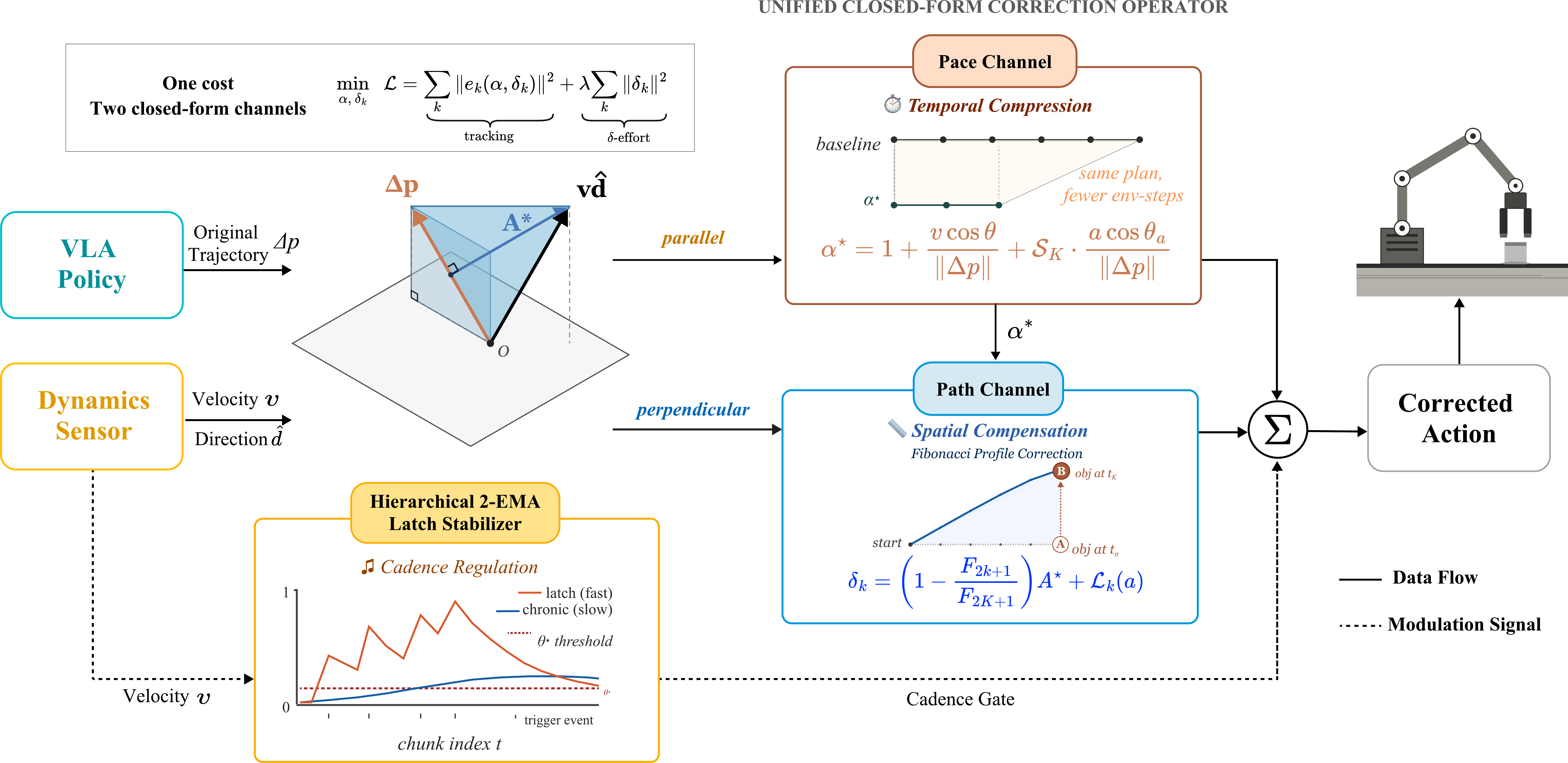}
    \caption{\textbf{Framework Overview.} Given a baseline action chunk $\Delta p$ from a frozen VLA policy and dynamics signals $(v, \hat{d})$ from the dynamics sensor, our framework minimizes a single quadratic cost over per-chunk tracking error and correction effort. Stationarity decomposes the optimum orthogonally into two closed-form channels: a \textit{Pace} Channel that absorbs the parallel component of $v\hat{d}$ as a temporal compression factor $\alpha^\star$, and a \textit{Path} Channel that absorbs the perpendicular residual $A^\star$ as a Fibonacci-profile spatial offset $\delta_k$. A \textit{hierarchical 2-EMA latch stabilizer} monitors the velocity stream and emits a Cadence Gate that caps the execution length when irregular regimes are detected. The summed correction yields the final corrected action with zero learnable parameters.}
    \vspace{-1em}
    \label{fig:framework}
\end{figure*}

\textbf{Vision-Language-Action Models.}
VLAs adapt pretrained vision-language backbones for robot control by mapping multimodal observations and language to action sequences~\citep{Brohan2023RT2VM,Kim2024OpenVLAAO,pi0}. Early designs decode actions autoregressively as discrete text tokens, enabling reuse of language-modeling objectives but limited by the resolution of binned actions and the cost of token-by-token decoding~\citep{Brohan2023RT2VM,Kim2024OpenVLAAO,Brohan2022RT1RT,Padalkar2023OpenXR}. More recent generalist policies attach a diffusion- or flow-matching action expert that emits continuous action chunks, recovering high-frequency control at the cost of grafting newly initialized weights onto the backbone~\citep{Chi2023DiffusionPV,pi0,Intelligence202505AV,Nvidia2025GR00TNA,Shukor2025SmolVLAAV,Liu2025HybridVLACD}. Across both lines, action chunking has emerged as the de facto control unit, where each inference call produces a fixed-horizon sequence executed open-loop before the next observation, trading reactivity for inference amortization~\citep{act,Liu2024RDT1BAD,Zhang2025ActionCA}.

\textbf{Dynamic Manipulation Benchmarks.} Robot manipulation benchmarks have largely standardized around static settings, with widely used VLA evaluation suites such as LIBERO~\citep{Liu2023LIBEROBK}, CALVIN~\citep{Mees2021CALVINAB}, ManiSkill~\citep{Mu2021ManiSkillGM,Gu2023ManiSkill2AU}, RoboCasa~\citep{Nasiriany2024RoboCasaLS}, and VLABench~\citep{Zhang2024VLABenchAL} measuring long-horizon planning, language grounding, or skill transfer while keeping objects stationary. Dynamic settings have only recently entered the VLA picture, primarily through DOM~\citep{Xie2026DynamicVLAAV} and DOMINO~\citep{Fang2026TowardsGR} as VLA-paired benchmarks targeting moving objects. These efforts establish that dynamic conditions degrade VLA performance, yet they treat motion as one axis intermixed with perception, generalization, or scene difficulty, and the underlying motion regimes are typically limited to uniform translation or simple acceleration~\citep{Zhang2025DynamicBC,BurgessLimerick2022DGBenchAO,Hassan2024GEMAG}. A controlled evaluation that varies motion alone, across uniform, accelerated, and irregular regimes, while holding tasks, objects, and scenes fixed remains an open need.

\textbf{Dynamics-Aware Vision-Language-Action Models.}
Existing remedies broadly follow two threads.
The first injects temporal or predictive cues into the backbone: FlowVLA~\citep{zhong2025flowvla}, PUMA~\citep{Fang2026TowardsGR}, and LaMP~\citep{Wang2026LaMPLV} feed historical optical or scene flow, TraceVLA~\citep{Zheng2024TraceVLAVT} overlays visual traces, MemoryVLA~\citep{Shi2025MemoryVLAPM} retrieves from episodic memory banks, and DreamVLA~\citep{Zhang2025DreamVLAAV}, WorldVLA~\citep{Cen2025WorldVLATA}, 4D-VLA~\citep{Zhang20254DVLASV}, FUTURE-VLA~\citep{Fan2026FUTUREVLAFU}, and SC-VLA~\citep{Liu2026SelfCorrectingVO} forecast future states through world models or predictive heads, all requiring retraining and architecture-specific integration~\citep{Fang2025RoboticVB}.
The second reduces inference latency while retaining the single-frame paradigm: DynamicVLA~\citep{Xie2026DynamicVLAAV} shrinks the backbone to 0.4B, PD-VLA~\citep{Song2025PDVLAAV} parallelizes autoregressive decoding, FASTer~\citep{Liu2025FASTerTE} compresses action tokenization, and others accelerate through token caching~\citep{Xu2025VLACacheEV,Tan2025ThinkTA}, discrete diffusion~\citep{liang2025discrete}, or asynchronous inference~\citep{zhang2024hirt}.
Orthogonal efforts repair chunk boundaries at inference time through temporal ensembling~\citep{act}, guided rejection sampling~\citep{Liu2024BidirectionalDI}, asynchronous inpainting~\citep{Black2025RealTimeEO,Black2025TrainingTimeAC}, learned correction heads~\citep{Sendai2025LeaveNO}, native continuation~\citep{Liu2026LearningNC}, or adaptive chunk sizing~\citep{Wen2026AdaptiveAC}, smoothing inter-chunk seams without addressing intra-chunk drift.

\section{Methodology}\label{method}

\subsection{Problem Formulation}\label{sec:problem}

A VLA policy maps an observation $o_t$ and a language instruction to an action chunk $A_t = (a_t, \ldots, a_{t+H-1})$, where each $a_k$ encodes an end-effector delta $\Delta p_k \in \mathbb{R}^3$ together with rotation and gripper commands. The robot executes the first $K \le H$ entries open-loop before re-querying the policy, with $T$ denoting the full chunk length. Let $\Delta p$ denote the representative per-step delta within this window, so the nominal trajectory is $p_k = k\Delta p$ for $k = 1, \ldots, K$. Absorbing the control timestep into \(v\), let \(v\hat d\) denote the target displacement per step along unit direction \(\hat d\). When the target moves during execution, the waypoints to track shift to \(\tilde p_k = k(\Delta p + v\hat d)\), while the chunk continues toward \(p_k\), yielding a tracking error \(\|p_k-\tilde p_k\|=kv\) that grows linearly with disturbance magnitude and step index and remains invisible to the policy until the next chunk is queried. To close this gap at inference time, we introduce a temporal-compression scalar $\alpha \ge 1$ and per-step spatial offsets $\{\delta_k\}_{k=0}^{K-1} \in \mathbb{R}^3$ on the chunk interior, so that the corrected delta at env-step $k$ becomes $u_k = \alpha \Delta p + \delta_k$. Introducing the residual disturbance $A := v\hat d - (\alpha - 1)\Delta p$ and the cumulative spatial offset $\sigma_k := \sum_{i=0}^{k-1}\delta_i$, the per-waypoint tracking error becomes
$$e_k(\alpha,\{\delta_i\}) \;=\; -kA + \sigma_k, \qquad k = 1, \ldots, K.$$
We then choose $(\alpha, \{\delta_k\})$ by minimizing
$$\min_{\alpha,\{\delta_k\}} \mathcal{L} \;=\; \sum_{k=1}^{K} \|e_k(\alpha,\{\delta_i\})\|^2 \;+\; \lambda \sum_{k=0}^{K-1} \|\delta_k\|^2,$$
balancing waypoint tracking against the effort of spatial deviation. This convex quadratic admits a closed-form minimizer whose two channels decompose orthogonally with respect to the disturbance direction, and the joint stationarity conditions yield
$$\Delta p \cdot \sum_{k=1}^{K} k\,e_k \;=\; 0, \qquad \lambda\,\delta_k + \sum_{j=k+1}^{K} e_j \;=\; 0.$$
We show next that the two correction degrees of freedom act on orthogonal subspaces, so the channels can be derived sequentially without loss of optimality.

\subsection{Pace Channel Correction}

Rotational invariance of the cost forces every $\delta_k$ at the optimum to inherit the direction of $A$, so $\sum_j j\,e_j$ lies parallel to $A$ and the first stationarity condition collapses to $\Delta p \cdot A = 0$. Expanding this orthogonality yields
$$\;\alpha^\star \;=\; 1 + \frac{v\cos\theta}{\|\Delta p\|}, \qquad \cos\theta \;=\; \hat d \cdot \widehat{\Delta p}.\;$$
The cosine factor ensures that only the disturbance component aligned with the plan modulates the pace, and substituting $\alpha^\star$ back into $A$ produces the orthogonal residual
$$A^\star \;=\; v\hat d - v\cos\theta\,\widehat{\Delta p} \;=\; v\hat d_\perp,$$
which lies entirely in the plane perpendicular to the planned direction. Geometrically, $\alpha^\star$ stretches the chunk's per-step magnitude exactly enough to keep the chunk endpoint $K\alpha^\star\Delta p$ aligned with the moving target's projection onto $\widehat{\Delta p}$, and the full wrapper reduces to the baseline VLA if and only if $v = 0$. At runtime, the compression is realized by setting $K_{\mathrm{exec}} = \max\!\bigl(K,\,\min(\lceil T/\alpha^\star \rceil,\,T)\bigr)$. Generalizing to an affine disturbance $v(t) = v_0 + at$ with possibly distinct directions $\hat d_v,\hat d_a$ ($\cos\theta_v = \hat d_v \cdot \widehat{\Delta p}$, $\cos\theta_a = \hat d_a \cdot \widehat{\Delta p}$) yields
$$\alpha^\star \;=\; 1 + \frac{v_0\cos\theta_v}{\|\Delta p\|} + \frac{3K(K{+}1)}{4(2K{+}1)}\cdot\frac{a\cos\theta_a}{\|\Delta p\|},$$
with the second-order coefficient scaling linearly in $K$, reflecting the longer integration window over which acceleration accumulates.

\subsection{Path Channel Correction}

The Path channel handles the residual $A^\star$, which cannot be absorbed by temporal scaling. Setting $\lambda = 1$ (generalized in Appendix~\ref{app:lambda}) and differencing the second stationarity condition in $k$ yields the 2D linear recurrence
$$\begin{pmatrix} \delta_{k+1}\\ e_{k+1} \end{pmatrix}
= \begin{pmatrix} 2 & 1\\ 1 & 1 \end{pmatrix}
\begin{pmatrix} \delta_{k}\\ e_{k} \end{pmatrix} - \begin{pmatrix} A^\star\\ A^\star \end{pmatrix}, \qquad e_0 = 0,\;\delta_K = 0.$$
The companion matrix has eigenvalues $\varphi^{\pm 2}$ where $\varphi = (1 + \sqrt 5)/2$ is the golden ratio. Solving the recursion under the boundary conditions and applying the identity $\varphi^n + \varphi^{-n} = \sqrt 5\,F_n$ for odd $n$ yields
$$\;\delta_k^\star \;=\; \left(1 - \frac{F_{2k+1}}{F_{2K+1}}\right) v\hat d_\perp, \qquad k = 0, \ldots, K-1,\;$$
where $F_n$ is the $n$-th Fibonacci number. The profile saturates from $\delta_0^\star \approx v\hat d_\perp$ at the chunk start to $\delta_{K-1}^\star \to \varphi^{-2} v\hat d_\perp \approx 0.618\,v\hat d_\perp$ as $K \to \infty$, with the boundary condition $\delta_K = 0$ ensuring the next chunk starts unbiased. This shape minimizes $\sum_k \|\delta_k\|^2$ while distributing the perpendicular displacement gradually across the executed window rather than concentrating it on any single env-step. Under the second-order disturbance, the same recurrence acquires an inhomogeneous term proportional to $B^\star := \tfrac{1}{2} a\hat d_{a,\perp}$, and linearity of the recurrence yields an additive decomposition into a Fibonacci first-order branch and a Lucas-polynomial second-order branch $\Lambda_k(K)$,
$$\delta_k^\star \;=\; \left(1 - \frac{F_{2k+1}}{F_{2K+1}}\right) A^\star + \Lambda_k(K)\,B^\star,$$
where the Lucas profile is the natural dual to Fibonacci on the same eigenvalue structure. Combined with $\alpha^\star$, the corrected delta $u_k = \alpha^\star\Delta p + \delta_k^\star$ is fully determined by the chunk geometry and the dynamics signal with no learnable parameter.

\subsection{Hierarchical 2-EMA Latch Stabilizer}

The closed forms above are exact under a quasi-stationary disturbance. Irregular regimes such as random walk, stop-and-go, and teleport violate this condition, and a single instantaneous reading of $v$ may briefly mislead $\alpha^\star$ into a long execution that the next observation will contradict. We complement the closed-form operator with a stateful regime classifier that detects sustained instability rather than reacting to single-step transients.

For each chunk reset at index $t$, the stabilizer reads only the velocity stream and computes a hard-thresholded direction-shift trigger $\tau_t = \mathbf{1}[\rho_{\mathrm{gt}}(t) < 1/2]$ from the cosine similarity $\rho_{\mathrm{gt}}(t) = \max(0,\,v_t\cdot v_{t-1} / (\|v_t\|\|v_{t-1}\|))$, which fires when the disturbance direction shifts beyond the natural midpoint. The stabilizer cascades a slow outer EMA with a fast inner EMA. The outer estimates the chronic trigger rate, $C_t = \beta_{\mathrm{out}}\tau_t + (1-\beta_{\mathrm{out}})C_{t-1}$, and feeds a Kalman-style sticky factor $s_t = C_t/(C_t + R_{\mathrm{TH}})$ that modulates the inner decay,
$$L_t \;=\; \begin{cases}
\beta_{\mathrm{in}} + (1-\beta_{\mathrm{in}})\,L_{t-1}, & \tau_t = 1,\\[0.3em]
\bigl[1 - \beta_{\mathrm{in}}(1 - s_t)\bigr]\,L_{t-1}, & \tau_t = 0.
\end{cases}$$
Under chronic instability ($s_t \to 1$) the inner state holds, while occasional triggers decay at the standard rate $\beta_{\mathrm{in}}$. The latch fires when $L_t$ exceeds a threshold and caps the executed chunk length under sustained irregularity (cadence gate),
$$m_t \;=\; \mathbf{1}[L_t > L_{\mathrm{th}}], \qquad K_{\mathrm{exec}} \;\le\; T/4 \text{ when } m_t = 1.$$
The latch admits a single free hyperparameter, the inner EMA rate $\beta_{\mathrm{in}}$, while the outer EMA rate $\beta_{\mathrm{out}} = 1 - 2^{-K/T}$ and the threshold $L_{\mathrm{th}} = R_{\mathrm{TH}} = \beta_{\mathrm{in}}(1-\beta_{\mathrm{in}})^2$ are derived from the chunk geometry by matching the outer EMA half-life to one chunk-budget cycle and calibrating $L_{\mathrm{th}}$ so that an isolated trigger sustains the latch for exactly two chunks.

\section{Experiments}\label{exp}

\begin{figure}[t]
  \centering
  \includegraphics[width=\linewidth]{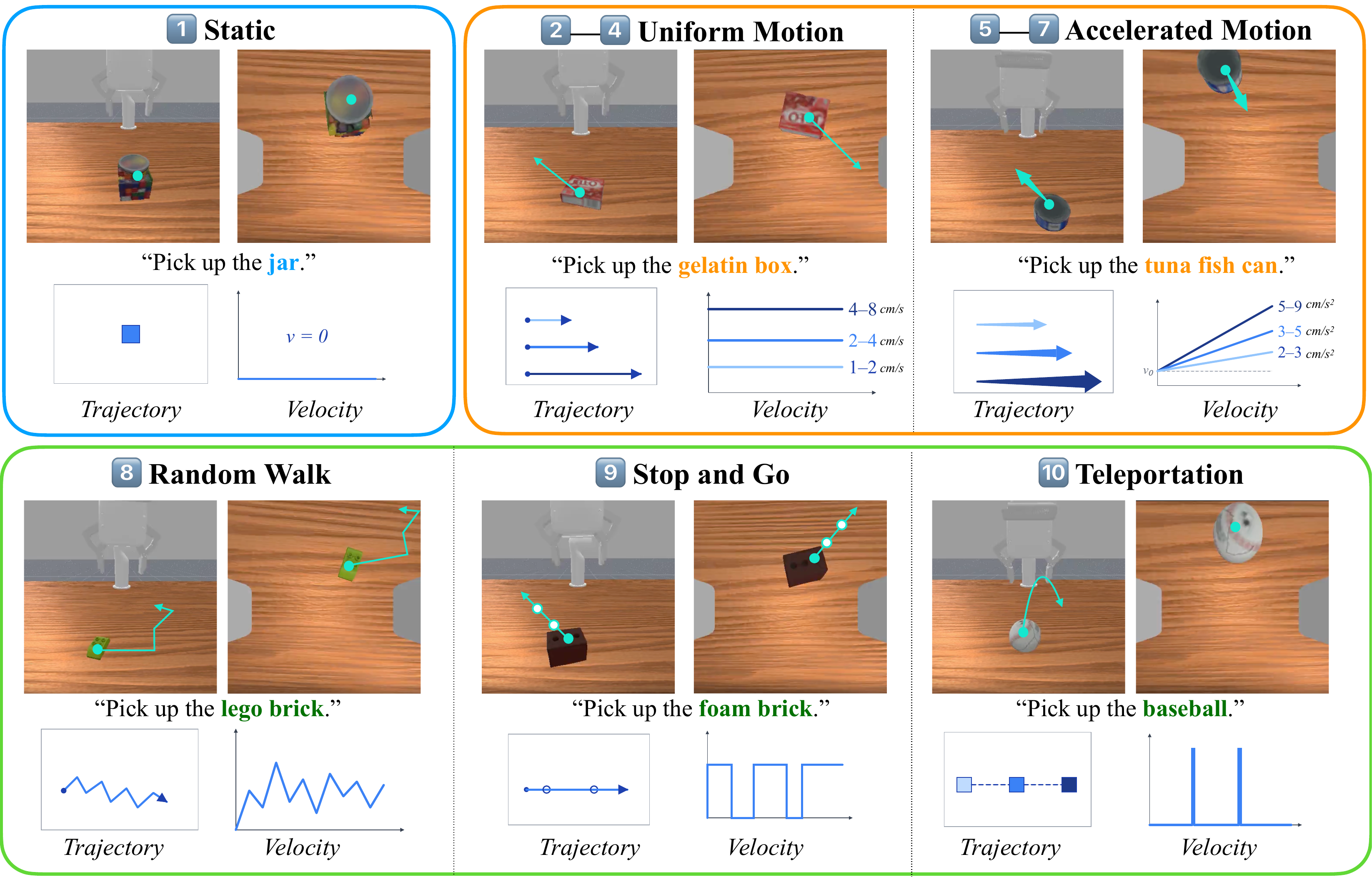}
  \caption{\textbf{\textsc{MoveBench} Overview.} \textsc{MoveBench} treats \textit{motion regimes} as the primary evaluation axis, comprising 10{,}000 trajectories ($\sim$460k frames) across 10 tasks with everyday household objects randomly sampled across regimes, spanning \textcolor{blue}{static}, \textcolor{orange}{regular}, and \textcolor{OliveGreen}{irregular} motion patterns at multiple difficulty levels. All non-motion factors are held identical, isolating motion as the sole variable.}
  \vspace{-1em}
  \label{fig:benchmark}
\end{figure}

\subsection{\textsc{MoveBench}}

We construct \textsc{MoveBench}, a benchmark for systematically studying how VLA models behave across environment-motion patterns. Built on ManiSkill with the SAPIEN engine and illustrated in Fig.~\ref{fig:benchmark}, \textsc{MoveBench} centers on a pick task in which an xArm6 grasps objects of varied shapes, with only the target's motion regime varying across all environments. The regimes form three families (uniform translation, accelerated motion, and irregular motion) plus a static control. Uniform and accelerated regimes are each graded over 3 difficulty levels (detailed in Appendix~\ref{app:movebench_details}). Higher difficulty shrinks the temporal window available to react. The irregular family covers three discrete event types (random walk, stop-and-go, and teleport), each at a single level, since they admit no continuous tunable scalar and instead probe regime-change response. Across the ten environments, each provides 1000 demonstrations, totaling 10K trajectories and $\sim$460K frames. By fixing the task, manipulator, and scene across environments, \textsc{MoveBench} isolates motion as the sole evaluation axis.

\subsection{Experimental Setup}

We compare PPC against 8 baselines spanning 2 categories. The first category covers state-of-the-art foundational VLAs and general-purpose visuomotor policies trained on large-scale robot data. The second category covers training-free inference-time wrappers for chunked-action execution improvements and dynamic-focused methods. PPC is integrated as an inference-time wrapper on top of four foundational backbones as illustrated in Table \ref{tab:baselines}, reusing each backbone's released checkpoint without any retraining or architectural modification, while all foundational baselines are fine-tuned on \textsc{MoveBench} demonstrations under their official recipes and dynamics-adaptive baselines follow their original deployment protocols. We choose $\pi_{0.5}$, the strongest foundational VLA, as ACT and BID's backbone for fairness. 100 trials are conducted for each task, resulting in 1,000 trials for each method. PPC's configuration is fixed throughout with $T = 16$, $K=2$, and Stabilizer EMA rate $\beta_{\mathrm{in}}=0.3$ (the single free knob), giving an inner-EMA half-life of ${\sim}\,2$ chunks under standard decay.

\subsection{Main Results}\label{main_resl}

\begin{table*}[t!]
\centering
\resizebox{\textwidth}{!}{%
\begin{tabular}{c l c ccc ccc ccc cc}
\toprule
\multirow{3}{*}{}
& \multirow{3}{*}{\textbf{Method}} 
& \multirow{3}{*}{\textbf{Static}} 
& \multicolumn{3}{c}{\textbf{Uniform Motion}} 
& \multicolumn{3}{c}{\textbf{Accelerated Motion}} 
& \multicolumn{3}{c}{\textbf{Irregular Motion}} 
& \multicolumn{2}{c}{\textbf{Average}} \\
\cmidrule(lr){4-6} \cmidrule(lr){7-9} \cmidrule(lr){10-12} \cmidrule(lr){13-14}
& & 
& \multicolumn{3}{c}{Moving}
& \multicolumn{3}{c}{Accelerating}
& \multirow{2}{*}{Rand.\,Walk} 
& \multirow{2}{*}{Stop\ \&\ Go} 
& \multirow{2}{*}{Teleport}
& \multirow{2}{*}{Dyn.\,Only} 
& \multirow{2}{*}{All} \\
\cmidrule(lr){4-6} \cmidrule(lr){7-9}
& & 
& Easy & Med. & Hard 
& Easy & Med. & Hard 
& & & & & \\
\midrule
\multirow{5}{*}{\rotatebox[origin=c]{90}{\textit{Found.}}}
& Diffusion Policy~\citep{Chi2023DiffusionPV} & 75 & 56 & 60 & 21 & 43 & 28 & 17 & 63 & 50 & 56 & $43.8_{\pm3.2}$ & $46.9_{\pm3.1}$ \\
& GR00T N1.6~\citep{Nvidia2025GR00TNA}      & \textbf{88} & 74 & 64 & 11 & 11 & 6  & 1  & 67 & 35 & \underline{67} & $37.3_{\pm3.2}$ & $42.4_{\pm3.1}$ \\
& SmolVLA~\citep{Shukor2025SmolVLAAV}          & 81 & 76 & 57 & 27 & 41 & 33 & 13 & 53 & 40 & 44 & $42.7_{\pm3.2}$ & $46.5_{\pm3.1}$ \\
& $\pi_0$~\citep{pi0}          & \underline{82} & 81 & 63 & 30 & 44 & 30 & 22 & 60 & 43 & 51 & $47.1_{\pm3.3}$ & $50.6_{\pm3.1}$ \\
& $\pi_{0.5}$~\citep{Intelligence202505AV}      & 80 & 85 & 78 & 34 & 58 & 43 & 29 & 54 & 48 & 60 & $54.3_{\pm3.3}$ & $56.9_{\pm3.1}$ \\
\midrule
\multirow{3}{*}{\rotatebox[origin=c]{90}{\textit{Comp.}}}
& ACT~\citep{act}         & \underline{82} & 79 & 77 & 19 & 69 & 50 & 30 & 53 & 48 & 1  & $47.3_{\pm3.3}$ & $50.8_{\pm3.1}$ \\
& BID~\citep{Liu2024BidirectionalDI}         & 79 & 80 & 75 & 29 & 57 & 50 & 33 & 68 & 51 & 48 & $54.6_{\pm3.3}$ & $57.0_{\pm3.1}$ \\
& DynamicVLA~\citep{Xie2026DynamicVLAAV}  & 70 & 73 & 57 & 20 & 45 & 42 & 29 & 49 & 40 & 24 & $42.1_{\pm3.2}$ & $44.9_{\pm3.1}$ \\
\midrule
\multirow{4}{*}{\rotatebox[origin=c]{90}{\textit{Ours}}}
& GR00T + PPC & $\textbf{88}^{*}$ 
  & \underline{86}\,\textcolor{green!60!black}{\scriptsize(+12\%)}
  & \underline{83}\,\textcolor{green!60!black}{\scriptsize(+19\%)}
  & 61\,\textcolor{green!60!black}{\scriptsize(+50\%)}
  & 70\,\textcolor{green!60!black}{\scriptsize(+59\%)}
  & 56\,\textcolor{green!60!black}{\scriptsize(+50\%)}
  & 33\,\textcolor{green!60!black}{\scriptsize(+32\%)}
  & \textbf{78}\,\textcolor{green!60!black}{\scriptsize(+11\%)}
  & 54\,\textcolor{green!60!black}{\scriptsize(+19\%)}
  & \textbf{74}\,\textcolor{green!60!black}{\scriptsize(+7\%)}
  & $66.1_{\pm3.1}$\,\textcolor{green!60!black}{\scriptsize(+28.8\%)}
  & $68.3_{\pm2.9}$\,\textcolor{green!60!black}{\scriptsize(+25.9\%)} \\
& SmolVLA + PPC & $81^{*}$ 
  & 69\,\textcolor{red}{\scriptsize($-$7\%)}
  & 69\,\textcolor{green!60!black}{\scriptsize(+12\%)}
  & 58\,\textcolor{green!60!black}{\scriptsize(+31\%)}
  & 58\,\textcolor{green!60!black}{\scriptsize(+17\%)}
  & 59\,\textcolor{green!60!black}{\scriptsize(+26\%)}
  & 35\,\textcolor{green!60!black}{\scriptsize(+22\%)}
  & 60\,\textcolor{green!60!black}{\scriptsize(+7\%)}
  & \textbf{71}\,\textcolor{green!60!black}{\scriptsize(+31\%)}
  & 53\,\textcolor{green!60!black}{\scriptsize(+9\%)}
  & $59.1_{\pm3.2}$\,\textcolor{green!60!black}{\scriptsize(+16.4\%)}
  & $61.3_{\pm3.0}$\,\textcolor{green!60!black}{\scriptsize(+14.8\%)} \\
& $\pi_0$ + PPC & $\underline{82}^{*}$ 
  & \underline{86}\,\textcolor{green!60!black}{\scriptsize(+5\%)}
  & 76\,\textcolor{green!60!black}{\scriptsize(+13\%)}
  & \underline{67}\,\textcolor{green!60!black}{\scriptsize(+37\%)}
  & \underline{73}\,\textcolor{green!60!black}{\scriptsize(+29\%)}
  & \underline{65}\,\textcolor{green!60!black}{\scriptsize(+35\%)}
  & \underline{57}\,\textcolor{green!60!black}{\scriptsize(+35\%)}
  & 71\,\textcolor{green!60!black}{\scriptsize(+11\%)}
  & \underline{67}\,\textcolor{green!60!black}{\scriptsize(+24\%)}
  & 52\,\textcolor{green!60!black}{\scriptsize(+1\%)}
  & $\underline{68.2}_{\pm3.0}$\,\textcolor{green!60!black}{\scriptsize(+21.1\%)}
  & $\underline{69.6}_{\pm2.9}$\,\textcolor{green!60!black}{\scriptsize(+19.0\%)} \\
& $\pi_{0.5}$ + PPC & $80^{*}$ 
  & \textbf{88}\,\textcolor{green!60!black}{\scriptsize(+3\%)}
  & \textbf{86}\,\textcolor{green!60!black}{\scriptsize(+8\%)}
  & \textbf{70}\,\textcolor{green!60!black}{\scriptsize(+36\%)}
  & \textbf{82}\,\textcolor{green!60!black}{\scriptsize(+24\%)}
  & \textbf{72}\,\textcolor{green!60!black}{\scriptsize(+29\%)}
  & \textbf{65}\,\textcolor{green!60!black}{\scriptsize(+36\%)}
  & \underline{74}\,\textcolor{green!60!black}{\scriptsize(+20\%)}
  & 66\,\textcolor{green!60!black}{\scriptsize(+18\%)}
  & 53\,\textcolor{red}{\scriptsize($-$7\%)}
  & $\textbf{72.9}_{\pm2.9}$\,\textcolor{green!60!black}{\scriptsize(+18.6\%)}
  & $\textbf{73.6}_{\pm2.7}$\,\textcolor{green!60!black}{\scriptsize(+16.7\%)} \\
\bottomrule
\end{tabular}%
}
\caption{\textbf{Success rate (\%) on \textsc{MoveBench} across all motion families.} PPC delivers consistent improvements over every state-of-the-art foundational VLA, and PPC-equipped VLAs surpass all comparison baselines in success rate. The static column for PPC variants is marked with $*$ to indicate that PPC defaults to the underlying VLA when no environmental motion is present. Subscript $\pm$ values on aggregate columns denote 95\% Clopper--Pearson confidence intervals over pooled trials.}
\vspace{-1.5em}
\label{tab:baselines}
\end{table*}

\begin{figure}[t!]
    \centering
    \begin{subfigure}[t]{0.40\textwidth}
        \centering
        \includegraphics[width=\textwidth]{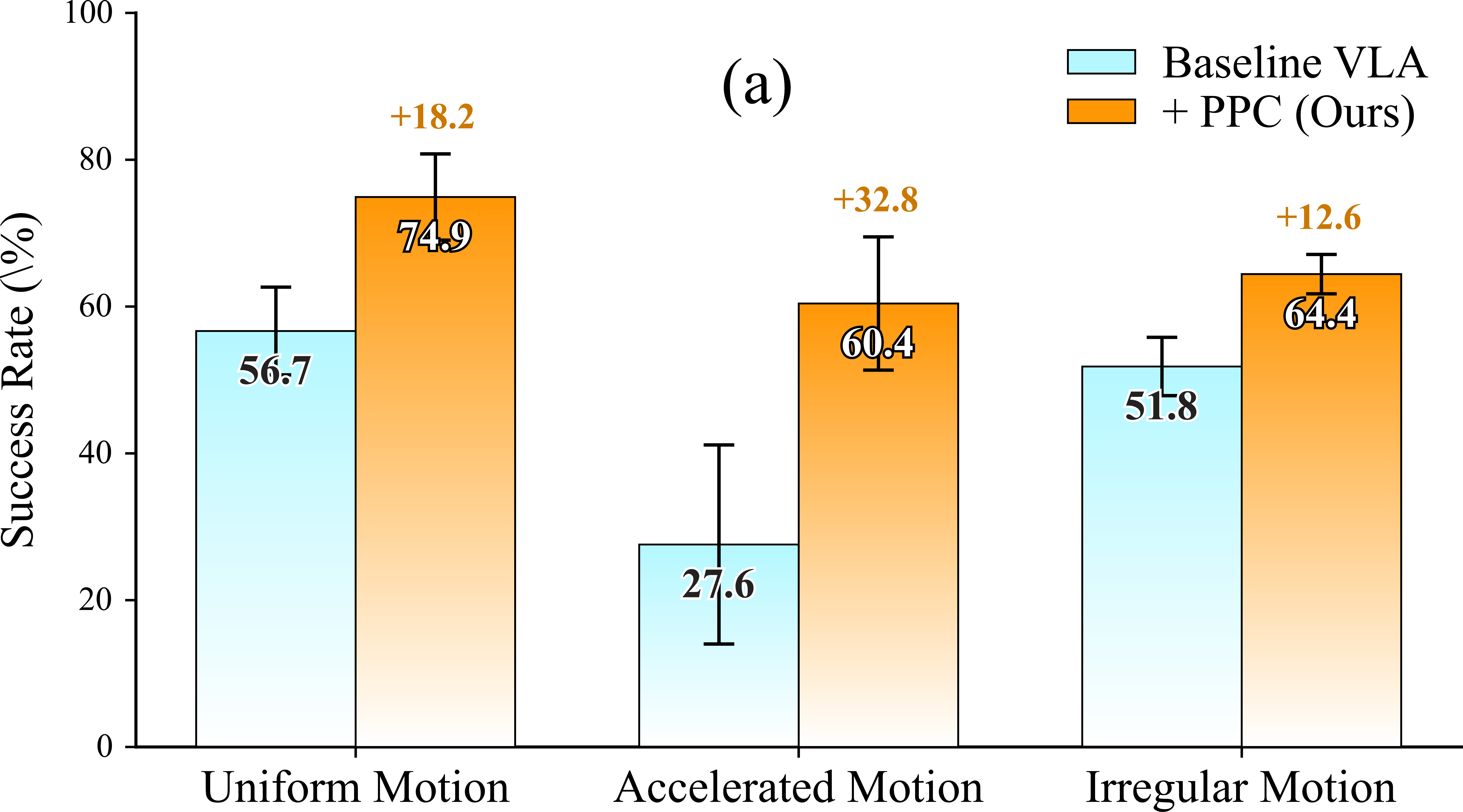}
        \label{fig:family_comparison}
    \end{subfigure}
    \hfill
    \begin{subfigure}[t]{0.50\textwidth}
        \centering
        \includegraphics[width=\textwidth]{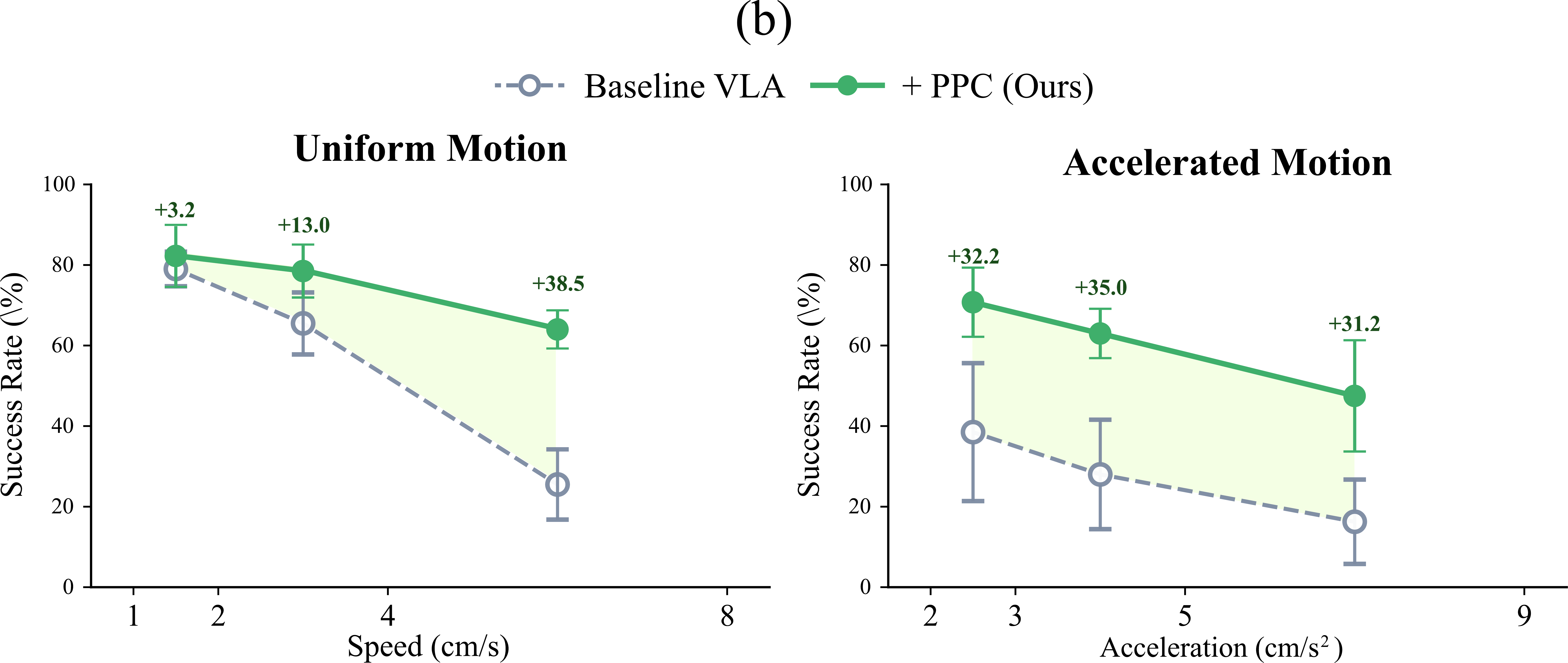}
    \end{subfigure}
    \caption{(a) Per-family success rate of baseline VLAs versus their PPC-equipped counterparts, averaged across four foundational backbones. PPC yields the largest gain on accelerated motion ($+32.8$). (b) Success rate as a function of physical speed and acceleration ranges. The gain from PPC grows monotonically with target speed in the uniform family while remaining consistently around $+30$ across the entire acceleration range.}
    \vspace{-2em}
    \label{fig:gain_analysis}
\end{figure}

Table~\ref{tab:baselines} reports success rates across all ten \textsc{MoveBench} environments. All foundational VLAs maintain strong static performance yet degrade sharply with increasing speed and acceleration, and neither chunk-level smoothing (ACT, BID) nor latency reduction (DynamicVLA) resolves this intra-chunk blindness. Three findings stand out. 

\textbf{PPC improves every foundational VLA across all motion families.} Wrapping the four foundational VLAs with PPC raises their dynamic-only average by $+16.4$ to $+28.8$ absolute points, with the best-equipped variant ($\pi_{0.5}$+PPC) reaching $72.9\%$ on dynamic environments and $73.6\%$ overall. Since $\alpha^\star\to 1$ and $\delta_k^\star\to 0$ when $v=0$, PPC degenerates to the identity by construction, preserving the full static capability without additional computation, while consistently improving performance across both regular and irregular motion regimes.

\textbf{The gain is largest where dynamics blindness is most severe.} Fig.~\ref{fig:gain_analysis} (a) shows that PPC yields its largest per-family improvement on accelerated motion ($+32.8$ averaged across backbones), followed by uniform ($+18.2$) and irregular ($+12.6$). This ordering directly reflects the closed-form structure. The Fibonacci-profile $\delta$ channel is designed to absorb the perpendicular residual that accumulates under sustained acceleration, explaining the largest gain in that family. Uniform motion is largely handled by the pace channel alone, while irregular regimes receive smaller but still positive gains as the latch-regulated cadence gate partially compensates for the weakened quasi-stationarity assumption. Fig.~\ref{fig:gain_analysis} (b) further reveals that the PPC gain grows monotonically with target speed in the uniform family (peaking at $+38.5$ at the hardest tier), while remaining consistently around $+30$ across the acceleration range, indicating that the second-order extension keeps pace with increasing acceleration.

\textbf{PPC-equipped VLAs surpass all comparison baselines.} Among the comparison methods, BID ($57.0\%$) and ACT ($50.8\%$) operate as inference-time wrappers on the same $\pi_{0.5}$ backbone yet fall short of every PPC variant, confirming that refining chunk outputs without an external dynamics signal cannot resolve intra-chunk blindness. ACT's near-zero teleport score ($1\%$) further demonstrates that a correction strategy mismatched to the motion regime can actively degrade performance below the uncorrected baseline, as temporal ensembling averages overlapping chunks so that a sudden object relocation causes stale actions to actively drag the end-effector toward the wrong position. DynamicVLA, despite being purpose-built for dynamic manipulation, underperforms even its backbone SmolVLA (further analyzed in Section~\ref{analysis}).

\subsection{Ablation Studies}

\begin{figure}[t!]
    \centering
    \begin{subfigure}[t]{0.48\textwidth}
        \centering
        \includegraphics[width=\textwidth, height=0.6\textwidth]{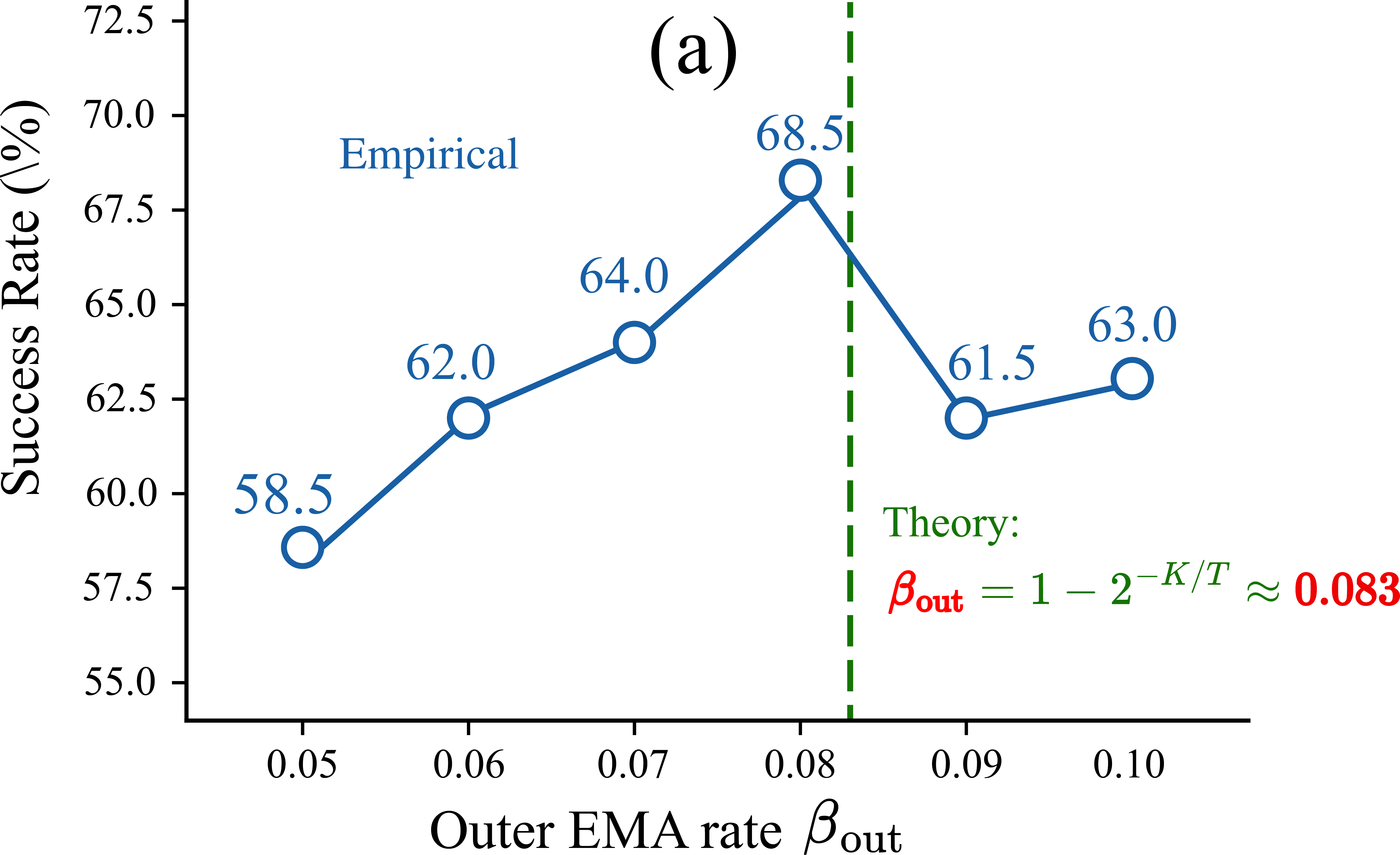}
    \end{subfigure}
    \hfill
    \begin{subfigure}[t]{0.48\textwidth}
        \centering
        \includegraphics[width=\textwidth, height=0.6\textwidth]{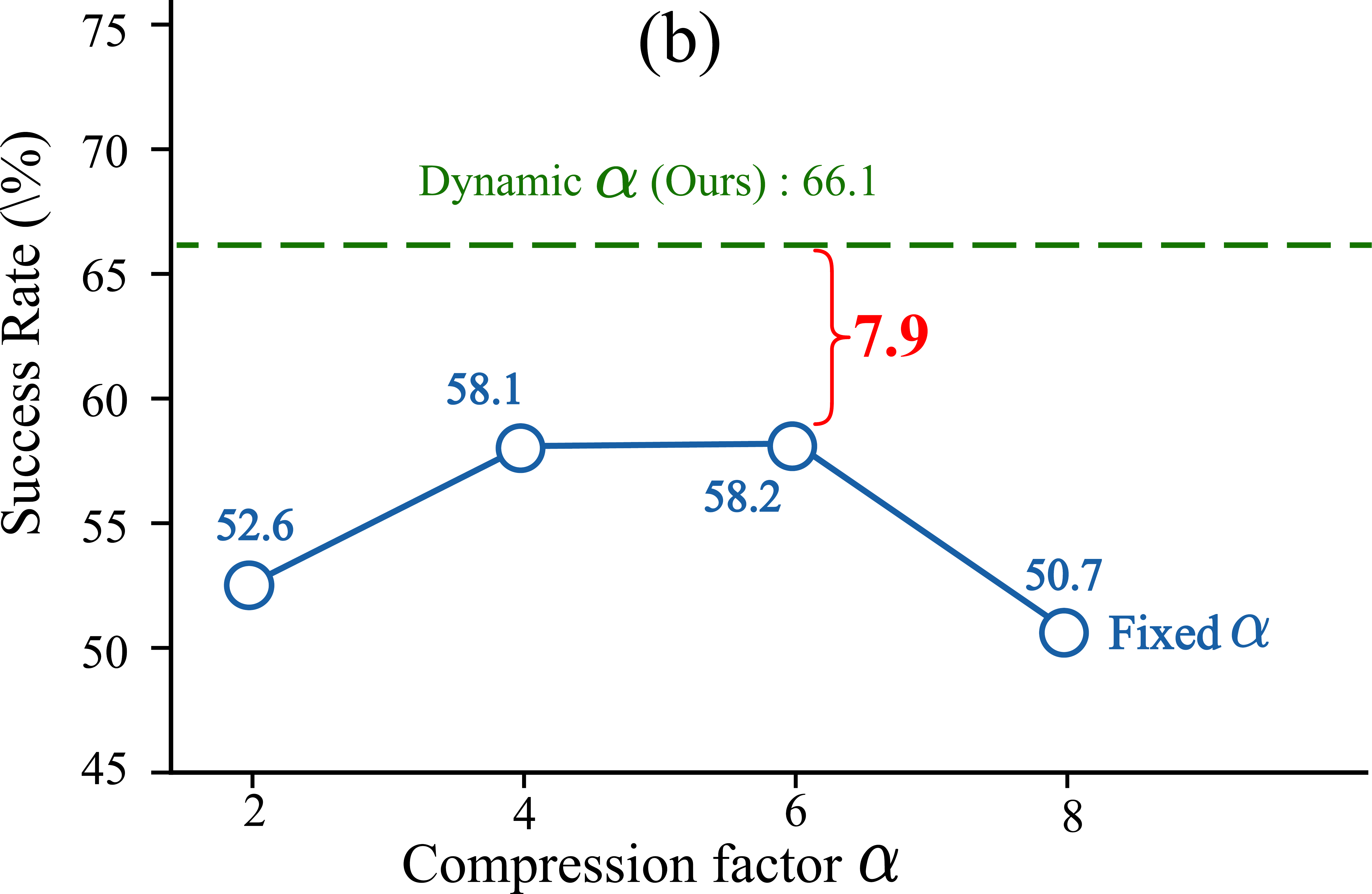}
    \end{subfigure}
    \caption{(a) Empirical sweep of $\beta_{\mathrm{out}}$ peaks at the closed-form theoretical value $\beta_{\mathrm{out}} = 1 - 2^{-K/T} \approx 0.083$, validating the latch derivation. (b) Dynamic $\alpha$ from the closed-form cost outperforms any fixed compression factor, confirming the necessity of per-chunk adaptive compression.}
    \label{fig:ablation_validations}
    \vspace{-1.5em}
\end{figure}

Comprehensive ablation is conducted to verify the effectiveness and robustness of PPC's components. All ablation experiments are performed on GR00T-N1.6+PPC across the dynamic environments of \textsc{MoveBench}, with 100 rollouts per environment, matching the setting in Section \ref{main_resl}.

\begin{wraptable}{r}{0.5\columnwidth}
\centering
\renewcommand{\arraystretch}{0.95}
\setlength{\tabcolsep}{3pt}
\footnotesize
\begin{tabular}{@{}l r r r r@{}}
\toprule
\textbf{Variant} & \textbf{Unif.} & \textbf{Accel.} & \textbf{Irreg.} & \textbf{Avg.} \\
\midrule
$\alpha$ w/o cos-$\theta$       & 68.7 \textcolor{red}{\tiny(-8.0)}  & 29.3 \textcolor{red}{\tiny(-23.7)} & 74.7 \textcolor{red}{\tiny(+6.0)}  & 57.6 \textcolor{red}{\tiny(-8.5)}  \\
Linear $\delta$ & 61.3 \textcolor{red}{\tiny(-15.4)} & 23.3 \textcolor{red}{\tiny(-29.7)} & 68.0 \textcolor{red}{\tiny(-0.7)}  & 50.9 \textcolor{red}{\tiny(-15.2)} \\
No $\alpha$ comp.  & 46.0 \textcolor{red}{\tiny(-30.7)} & 22.3 \textcolor{red}{\tiny(-30.7)} & 45.0 \textcolor{red}{\tiny(-23.7)} & 37.8 \textcolor{red}{\tiny(-28.3)} \\
$\delta$ w/o $\perp$ proj. & 68.7 \textcolor{red}{\tiny(-8.0)}  & 46.0 \textcolor{red}{\tiny(-7.0)}  & 64.3 \textcolor{red}{\tiny(-4.4)}  & 59.7 \textcolor{red}{\tiny(-6.4)}  \\
No $\delta$ offsets  & 52.0 \textcolor{red}{\tiny(-24.7)} & 9.3 \textcolor{red}{\tiny(-43.7)}  & 64.3 \textcolor{red}{\tiny(-4.4)}  & 41.9 \textcolor{red}{\tiny(-24.2)} \\
\midrule
\textbf{PPC (full)} & \textbf{76.7}  & \textbf{53.0}  & \textbf{68.7}  & \textbf{66.1}  \\
\bottomrule
\end{tabular}
\caption{\textbf{Closed-form structural ablations.} Removing or modifying any of the closed-form components consistently degrades performance, confirming that all design choices are necessary.}
\label{tab:ablation_structural}
 \vspace{-1.5em}
\end{wraptable}

\textbf{Closed-form structural ablations.} As shown in Table \ref{tab:ablation_structural}, all closed-form components are necessary, with every ablation falling below full PPC's $66.1\%$ overall success. Removing the $\alpha$ compression channel causes the largest collapse ($-28.3$ points), with near-uniform losses across all three motion families, confirming $\alpha$ as the dominant correction mechanism. Removing the $\delta$ offsets channel costs $24.2$ points overall but the loss is highly concentrated on accelerated motion (single-digit success) while irregular regimes are barely affected, indicating that $\alpha$ corrects globally while $\delta$ specifically absorbs the perpendicular drift accumulating under sustained acceleration. Removing the cos-$\theta$ projection in $\alpha$ shows the same directional split, hurting accelerated motion but slightly helping irregular regimes, since the unprojected formula yields a larger $\alpha$ that overshoots under consistent directional motion yet aids reactivity under rapidly-shifting directions. The Fibonacci profile and the $\perp$ projection on $\delta$ contribute smaller but consistent gains, acting as shape-level refinements within the $\delta$ channel.

\textbf{EMA Stabilizer Ablation.} Disabling the latch costs $6.0$ points overall (Uniform $74.0_{\textcolor{red}{\downarrow 2.7}}$, Accel $45.3_{\textcolor{red}{\downarrow 7.7}}$, Irregular $60.7_{\textcolor{red}{\downarrow 8.0}}$, Avg $60.0_{\textcolor{red}{\downarrow 6.1}}$), with the loss concentrated on irregular regimes and minimal on uniform motion. This asymmetry matches the latch's role as a regime-instability detector, activating only when the closed-form's quasi-stationarity assumption breaks down.

\textbf{$2^{nd}$-Order Channel Ablation.} Removing the Lucas branch costs $3.4$ points overall (Uniform $76.3_{\textcolor{red}{\downarrow 0.4}}$, Accel $45.7_{\textcolor{red}{\downarrow 7.3}}$, Irregular $66.0_{\textcolor{red}{\downarrow 2.7}}$, Avg $62.7_{\textcolor{red}{\downarrow 3.4}}$), with the loss concentrated on accelerated motion and negligible on uniform regimes, matching the Lucas branch's role as a $2^{nd}$-order corrector.

\begin{figure}[t]
  \centering
  \begin{subfigure}[b]{0.45\textwidth}
    \centering
    \includegraphics[width=\textwidth]{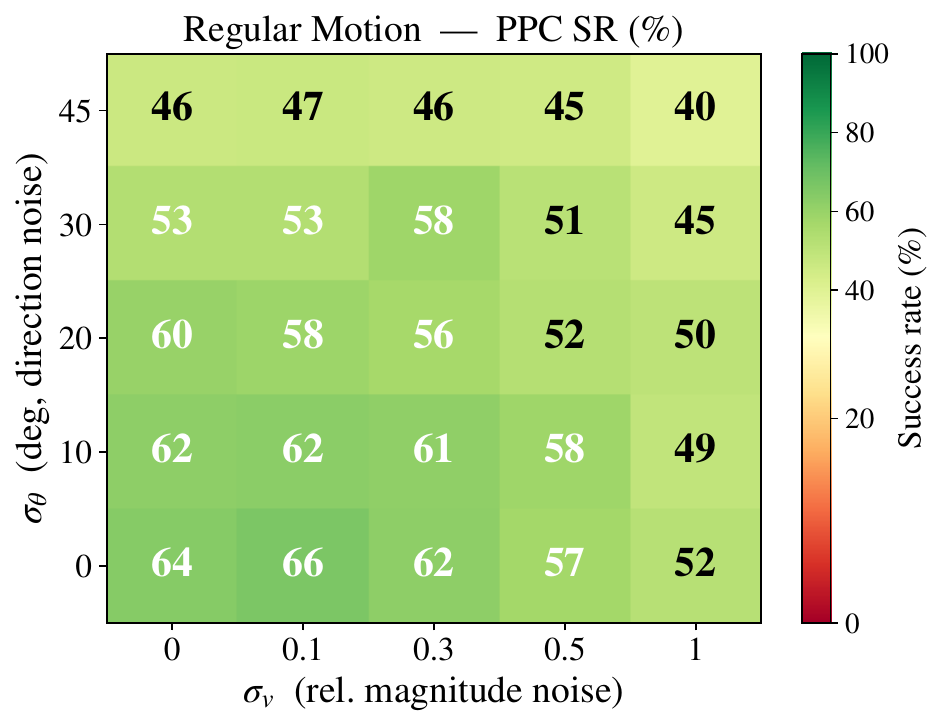}
    \label{fig:noise-accelhard}
  \end{subfigure}
  \hfill
  \begin{subfigure}[b]{0.45\textwidth}
    \centering
    \includegraphics[width=\textwidth]{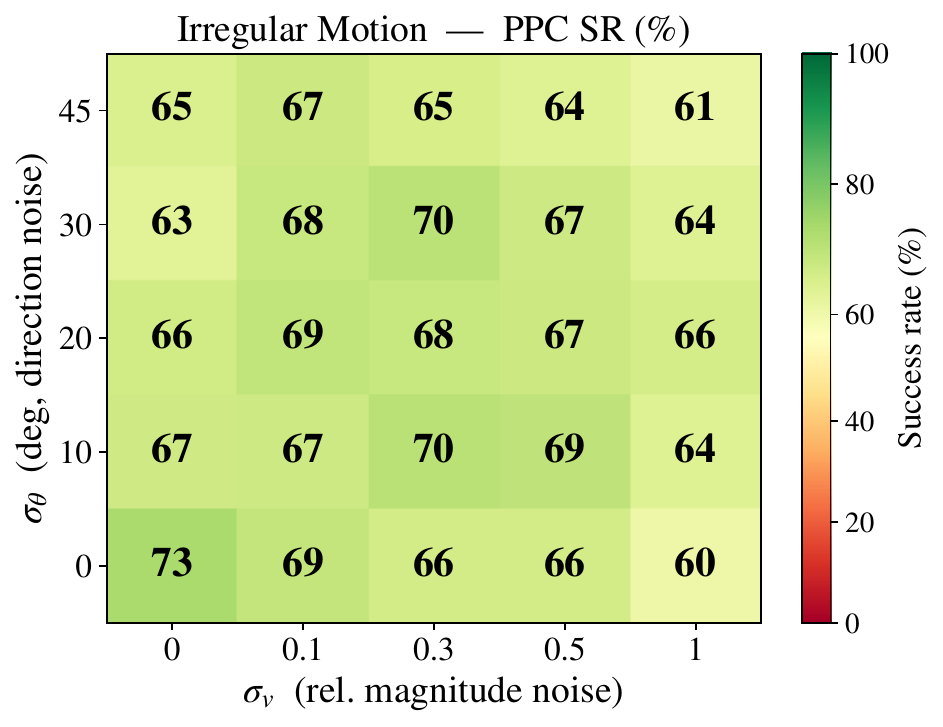}
    \label{fig:noise-randomwalk}
  \end{subfigure}
  \vspace{-1.5em}
  \caption{\textbf{Robustness to perception noise.} Success rate (\%) under varying magnitude noise $\sigma_v$ and directional noise $\sigma_\theta$ on the velocity signal. PPC remains above the bare baseline across all conditions.}
  \vspace{-1.5em}
  \label{fig:noise-robustness}
\end{figure}

\textbf{$\boldsymbol{\beta_{\text{outer}}}.$ theory validation} As illustrated in Fig.\ref{fig:ablation_validations} (a), sweeping $\beta_\mathrm{out}$ on irregular regimes (rand. walk and stop \& go) yields a peak success rate of $68\%$ at $\beta_\mathrm{out} \approx 0.08$, which closely matches the theoretical value $1 - 2^{-K/T} \approx 0.083$ derived in Section~\ref{method}. This empirical-theoretical alignment validates the closed-form derivation, confirming that $\beta_\mathrm{in}$ is the only true free hyperparameter.

\textbf{Dynamic vs. fix $\boldsymbol{\alpha}$.} As illustrated in Fig.\ref{fig:ablation_validations} (b), the dynamic $\alpha$ derived from the closed-form cost reaches $66.1\%$ overall, surpassing every fixed setting by a clear margin. Among the static alternatives, $\alpha=4$ and $\alpha=6$ peak at around $58\%$ and degrade on either side. The collapse at $\alpha=8$ confirms that blindly accelerating execution catches up to target speed at the cost of fine-grained control, where only the per-chunk adaptive $\alpha$ from the closed-form derivation balances the two.

\textbf{Robustness to perception noise} As illustrated in Fig.~\ref{fig:noise-robustness}, we inject synthetic magnitude noise $\sigma_v$ and directional noise $\sigma_\theta$ to simulate the error of real-world tracking hardware. Across all noise conditions, averaged over the six regular-motion environments and the three irregular-motion environments, PPC never falls below the corresponding bare-backbone baseline. Under noise levels typical of depth-camera and visual-tracker pipelines ($\sigma_v \leq 0.3$, $\sigma_\theta \leq 20°$), the correction operator retains the majority of its oracle-signal gains, confirming robustness sufficient for potential physical deployment.

\subsection{Analysis}\label{analysis}

\textbf{Latency is not the only bottleneck for dynamic manipulation.} DynamicVLA achieves the lowest inference latency among all methods via its compact 0.4B architecture, yet scores only $44.9\%$ on \textsc{MoveBench}, below even the much slower $\pi_{0.5}$ ($56.9\%$). This inversion shows that the dominant failure mode is not how often the policy re-plans, but that each chunk remains blind to motion during its execution window. Its static score ($70\%$) also regresses from its backbone SmolVLA ($81\%$), confirming that indiscriminate high-frequency re-inference disrupts inter-chunk coherence even absent any dynamic demand.

\textbf{Motion regime matters more than motion speed.} Across all foundational VLAs, accelerated motion causes the steepest performance collapse despite its physical speed often being lower than the hardest uniform tier, because acceleration accumulates drift nonlinearly within a chunk. Irregular regimes further reveal that even moderate-speed motion becomes challenging when its direction is unpredictable. These patterns would be invisible in benchmarks that only vary speed along a single motion type, validating \textsc{MoveBench}'s design of isolating regime as a first-class evaluation axis.

\textbf{Intra-chunk compensation scales with difficulty.}
Fig.~\ref{fig:gain_analysis} shows that PPC's gain over the bare backbone grows monotonically with target speed, reaching $+38.5$ at the hardest uniform tier, and remains consistently around $+30$ across the entire acceleration range. This scaling confirms that the closed-form correction absorbs progressively larger disturbances without saturating, and that explicit intra-chunk compensation is the effective lever for dynamic manipulation.

\section{Conclusion}

We present Pace-and-Path Correction (PPC), a closed-form, training-free, inference-time wrapper that explicitly compensates for environment dynamics in chunked-action VLAs. By decomposing the per-chunk correction objective into orthogonal pace and path channels, PPC introduces no learnable parameters and remains backbone-agnostic, deployable on top of any released VLA without retraining or architectural modification. We further introduce \textsc{MoveBench}, a benchmark that isolates motion regime as the sole evaluation axis for systematically studying chunked-VLA behavior under diverse motion patterns. Extensive experiments demonstrate that PPC consistently improves foundational VLAs and outperforms state-of-the-art dynamic-adaptive methods. Future work includes validating PPC with learned tracking pipelines and extending the formulation to multi-object dynamic scenes and manipulation primitives beyond pick.

\bibliographystyle{unsrtnat}
\bibliography{references}

\begin{thebibliography}{53}
\providecommand{\natexlab}[1]{#1}
\providecommand{\url}[1]{\texttt{#1}}
\expandafter\ifx\csname urlstyle\endcsname\relax
  \providecommand{\doi}[1]{doi: #1}\else
  \providecommand{\doi}{doi: \begingroup \urlstyle{rm}\Url}\fi

\bibitem[Zhang et~al.(2025{\natexlab{a}})Zhang, Wang, and Chen]{Zhang2025DynamicBC}
Yifan Zhang, Ruiping Wang, and Xilin Chen.
\newblock Dynamic behavior cloning with temporal feature prediction: Enhancing robotic arm manipulation in moving object tasks.
\newblock \emph{IEEE Robotics and Automation Letters}, 10:\penalty0 5209--5216, 2025{\natexlab{a}}.

\bibitem[Fang et~al.(2026)Fang, Li, Wang, Xi, Liang, and Bai]{Fang2026TowardsGR}
Heng Fang, Shangru Li, Shuhang Wang, Xuan Xi, Dingkang Liang, and Xiang Bai.
\newblock Towards generalizable robotic manipulation in dynamic environments.
\newblock 2026.

\bibitem[Xie et~al.(2026)Xie, Wen, Zheng, Chen, Hong, Diao, and Liu]{Xie2026DynamicVLAAV}
Haozhe Xie, Beichen Wen, Jia Zheng, Zhaoxi Chen, Fangzhou Hong, Haiwen Diao, and Ziwei Liu.
\newblock Dynamicvla: A vision-language-action model for dynamic object manipulation.
\newblock \emph{ArXiv}, abs/2601.22153, 2026.

\bibitem[Hu et~al.(2023)Hu, Xie, Jain, Francis, Patrikar, Keetha, Kim, Xie, Zhang, Zhao, Chong, Wang, Sycara, Johnson-Roberson, Batra, Wang, Scherer, Kira, Xia, and Bisk]{Hu2023TowardGR}
Yafei Hu, Quanting Xie, Vidhi Jain, Jonathan Francis, Jay Patrikar, Nikhil~Varma Keetha, Seungchan Kim, Yaqi Xie, Tianyi Zhang, Shibo Zhao, Yu~Quan Chong, Chen Wang, Katia~P. Sycara, Matthew Johnson-Roberson, Dhruv Batra, Xiaolong Wang, Sebastian Scherer, Zsolt Kira, Fei Xia, and Yonatan Bisk.
\newblock Toward general-purpose robots via foundation models: A survey and meta-analysis.
\newblock \emph{ArXiv}, abs/2312.08782, 2023.

\bibitem[Ma et~al.(2024)Ma, Song, Zhuang, Hao, and King]{Ma2024ASO}
Yueen Ma, Zixing Song, Yuzheng Zhuang, Jianye Hao, and Irwin King.
\newblock A survey on vision-language-action models for embodied ai.
\newblock \emph{ArXiv}, abs/2405.14093, 2024.

\bibitem[Team et~al.(2024)Team, Ghosh, Walke, Pertsch, Black, Mees, Dasari, Hejna, Kreiman, Xu, Luo, Tan, Sanketi, Vuong, Xiao, Sadigh, Finn, and Levine]{Team2024OctoAO}
Octo~Model Team, Dibya Ghosh, Homer~Rich Walke, Karl Pertsch, Kevin Black, Oier Mees, Sudeep Dasari, Joey Hejna, Tobias Kreiman, Charles Xu, Jianlan Luo, You~Liang Tan, Pannag~R. Sanketi, Quan Vuong, Ted Xiao, Dorsa Sadigh, Chelsea Finn, and Sergey Levine.
\newblock Octo: An open-source generalist robot policy.
\newblock \emph{ArXiv}, abs/2405.12213, 2024.

\bibitem[Brohan et~al.(2023)Brohan, Brown, Carbajal, Chebotar, Choromanski, Ding, Driess, Dubey, Finn, Florence, Fu, Arenas, Gopalakrishnan, Han, Hausman, Herzog, Hsu, Ichter, Irpan, Joshi, Julian, Kalashnikov, Kuang, Leal, Levine, Michalewski, Mordatch, Pertsch, Rao, Reymann, Ryoo, Salazar, Sanketi, Sermanet, Singh, Singh, Soricut, Tran, Vanhoucke, Vuong, Wahid, Welker, Wohlhart, Xiao, Yu, and Zitkovich]{Brohan2023RT2VM}
Anthony Brohan, Noah Brown, Justice Carbajal, Yevgen Chebotar, Krzysztof Choromanski, Tianli Ding, Danny Driess, Kumar~Avinava Dubey, Chelsea Finn, Peter~R. Florence, Chuyuan Fu, Montse~Gonzalez Arenas, Keerthana Gopalakrishnan, Kehang Han, Karol Hausman, Alexander Herzog, Jasmine Hsu, Brian Ichter, Alex Irpan, Nikhil~J. Joshi, Ryan~C. Julian, Dmitry Kalashnikov, Yuheng Kuang, Isabel Leal, Sergey Levine, Henryk Michalewski, Igor Mordatch, Karl Pertsch, Kanishka Rao, Krista Reymann, Michael~S. Ryoo, Grecia Salazar, Pannag~R. Sanketi, Pierre Sermanet, Jaspiar Singh, Anikait Singh, Radu Soricut, Huong Tran, Vincent Vanhoucke, Quan~Ho Vuong, Ayzaan Wahid, Stefan Welker, Paul Wohlhart, Ted Xiao, Tianhe Yu, and Brianna Zitkovich.
\newblock Rt-2: Vision-language-action models transfer web knowledge to robotic control.
\newblock \emph{ArXiv}, abs/2307.15818, 2023.

\bibitem[Kim et~al.(2024)Kim, Pertsch, Karamcheti, Xiao, Balakrishna, Nair, Rafailov, Foster, Lam, Sanketi, Vuong, Kollar, Burchfiel, Tedrake, Sadigh, Levine, Liang, and Finn]{Kim2024OpenVLAAO}
Moo~Jin Kim, Karl Pertsch, Siddharth Karamcheti, Ted Xiao, Ashwin Balakrishna, Suraj Nair, Rafael Rafailov, Ethan~Paul Foster, Grace Lam, Pannag~R. Sanketi, Quan Vuong, Thomas Kollar, Benjamin Burchfiel, Russ Tedrake, Dorsa Sadigh, Sergey Levine, Percy Liang, and Chelsea Finn.
\newblock Openvla: An open-source vision-language-action model.
\newblock \emph{ArXiv}, abs/2406.09246, 2024.

\bibitem[Black et~al.(2024)Black, Brown, Driess, Esmail, Equi, Finn, Fusai, Groom, Hausman, Ichter, Jakubczak, Jones, Ke, Levine, Li-Bell, Mothukuri, Nair, Pertsch, Shi, Tanner, Vuong, Walling, Wang, and Zhilinsky]{pi0}
Kevin Black, Noah Brown, Danny Driess, Adnan Esmail, Michael Equi, Chelsea Finn, Niccolo Fusai, Lachy Groom, Karol Hausman, Brian Ichter, Szymon Jakubczak, Tim Jones, Liyiming Ke, Sergey Levine, Adrian Li-Bell, Mohith Mothukuri, Suraj Nair, Karl Pertsch, Lucy~Xiaoyang Shi, James Tanner, Quan Vuong, Anna Walling, Haohuan Wang, and Ury Zhilinsky.
\newblock $\pi$0: A vision-language-action flow model for general robot control.
\newblock \emph{ArXiv}, abs/2410.24164, 2024.

\bibitem[Zhao et~al.(2023)Zhao, Kumar, Levine, and Finn]{act}
Tony Zhao, Vikash Kumar, Sergey Levine, and Chelsea Finn.
\newblock Learning fine-grained bimanual manipulation with low-cost hardware.
\newblock \emph{ArXiv}, abs/2304.13705, 2023.

\bibitem[Chi et~al.(2023)Chi, Feng, Du, Xu, Cousineau, Burchfiel, and Song]{Chi2023DiffusionPV}
Cheng Chi, Siyuan Feng, Yilun Du, Zhenjia Xu, Eric Cousineau, Benjamin Burchfiel, and Shuran Song.
\newblock Diffusion policy: Visuomotor policy learning via action diffusion.
\newblock \emph{The International Journal of Robotics Research}, 44:\penalty0 1684 -- 1704, 2023.

\bibitem[Nvidia et~al.(2025)Nvidia, Bjorck, Casta{\~n}eda, Cherniadev, Da, Ding, LinxiJimFan, Fang, Fox, Hu, Huang, Jang, Jiang, Kautz, Kundalia, Lao, Li, Lin, Lin, Liu, Llontop, Magne, Mandlekar, Narayan, Nasiriany, Reed, Tan, Wang, Wang, Wang, Wang, Xiang, Xie, Xu, Xu, Ye, Yu, Zhang, Zhang, Zhao, Zheng, and Zhu]{Nvidia2025GR00TNA}
Nvidia, Johan Bjorck, Fernando Casta{\~n}eda, Nikita Cherniadev, Xingye Da, Runyu Ding, LinxiJimFan, Yu~Fang, Dieter Fox, Fengyuan Hu, Spencer Huang, Joel Jang, Zhenyuan Jiang, Jan Kautz, Kaushil Kundalia, Lawrence Lao, Zhiqi Li, Zongyu Lin, Kevin Lin, Guilin Liu, Edith Llontop, Loic Magne, Ajay Mandlekar, Avnish Narayan, Soroush Nasiriany, Scott Reed, You~Liang Tan, Guanzhi Wang, Zu~Wang, Jing Wang, Qi~Wang, Jiannan Xiang, Yuqi Xie, Yinzhen Xu, Zhen-Teng Xu, Seonghyeon Ye, Zhiding Yu, Ao~Zhang, Hao Zhang, Yizhou Zhao, Ruijie Zheng, and Yuke Zhu.
\newblock Gr00t n1: An open foundation model for generalist humanoid robots.
\newblock \emph{ArXiv}, abs/2503.14734, 2025.

\bibitem[Liu et~al.(2023)Liu, Zhu, Gao, Feng, Liu, Zhu, and Stone]{Liu2023LIBEROBK}
Bo~Liu, Yifeng Zhu, Chongkai Gao, Yihao Feng, Qian Liu, Yuke Zhu, and Peter Stone.
\newblock Libero: Benchmarking knowledge transfer for lifelong robot learning.
\newblock \emph{ArXiv}, abs/2306.03310, 2023.

\bibitem[Mees et~al.(2021)Mees, Hermann, Rosete-Beas, and Burgard]{Mees2021CALVINAB}
Oier Mees, Luk{\'a}s Hermann, Erick Rosete-Beas, and Wolfram Burgard.
\newblock Calvin: A benchmark for language-conditioned policy learning for long-horizon robot manipulation tasks.
\newblock \emph{IEEE Robotics and Automation Letters}, 7:\penalty0 7327--7334, 2021.

\bibitem[Zhong et~al.(2025)Zhong, Yan, Li, Liu, Gong, Zhang, Song, Chen, Zheng, Wang, and Li]{zhong2025flowvla}
Zhide Zhong, Haodong Yan, Junfeng Li, Xiangcheng Liu, Xin Gong, Tianran Zhang, Wenxuan Song, Jiayi Chen, Xinhu Zheng, Hesheng Wang, and Haoang Li.
\newblock Flowvla: Visual chain of thought-based motion reasoning for vision-language-action models.
\newblock \emph{arXiv preprint arXiv:2508.18269}, 2025.

\bibitem[Fang et~al.(2025)Fang, Ranasinghe, Xue, Zhou, Tan, Xu, Heinecke, Xiong, Savarese, Szafir, Ding, Ryoo, and Niebles]{Fang2025RoboticVB}
Yu~Fang, Kanchana Ranasinghe, Le~Xue, Honglu Zhou, Juntao Tan, Ran Xu, Shelby Heinecke, Caiming Xiong, Silvio Savarese, Danielle~Albers Szafir, Mingyu Ding, Michael~S. Ryoo, and Juan~Carlos Niebles.
\newblock Robotic vla benefits from joint learning with motion image diffusion.
\newblock \emph{ArXiv}, abs/2512.18007, 2025.

\bibitem[Zheng et~al.(2024)Zheng, Liang, Huang, Gao, Daum'e, Kolobov, Huang, and Yang]{Zheng2024TraceVLAVT}
Ruijie Zheng, Yongyuan Liang, Shuaiyi Huang, Jianfeng Gao, Hal Daum'e, Andrey Kolobov, Furong Huang, and Jianwei Yang.
\newblock Tracevla: Visual trace prompting enhances spatial-temporal awareness for generalist robotic policies.
\newblock \emph{ArXiv}, abs/2412.10345, 2024.

\bibitem[Shi et~al.(2025)Shi, Xie, Liu, Sun, Liu, Wang, Zhou, Fan, Zhang, and Huang]{Shi2025MemoryVLAPM}
Hao Shi, Bin Xie, Yingfei Liu, Lin Sun, Feng Liu, Tiancai Wang, Erjin Zhou, Haoqiang Fan, Xiangyu Zhang, and Gao Huang.
\newblock Memoryvla: Perceptual-cognitive memory in vision-language-action models for robotic manipulation.
\newblock \emph{ArXiv}, abs/2508.19236, 2025.

\bibitem[Zhang et~al.(2025{\natexlab{b}})Zhang, Liu, Qi, Wang, Yu, Zhang, Dong, He, Wang, Zhang, Yi, Zeng, and Jin]{Zhang2025DreamVLAAV}
Wenyao Zhang, Hongsi Liu, Zekun Qi, Yunnan Wang, Xinqiang Yu, Jiazhao Zhang, Runpei Dong, Jiawei He, He~Wang, Zhizheng Zhang, Li~Yi, Wenjun Zeng, and Xin Jin.
\newblock Dreamvla: A vision-language-action model dreamed with comprehensive world knowledge.
\newblock \emph{ArXiv}, abs/2507.04447, 2025{\natexlab{b}}.

\bibitem[Cen et~al.(2025)Cen, Yu, Yuan, Jiang, Huang, Guo, Li, Song, Luo, Wang, Zhao, and Chen]{Cen2025WorldVLATA}
Jun Cen, Chaohui Yu, Hangjie Yuan, Yuming Jiang, Siteng Huang, Jiayan Guo, Xin Li, Yibing Song, Hao Luo, Fan Wang, Deli Zhao, and Hao Chen.
\newblock Worldvla: Towards autoregressive action world model.
\newblock \emph{ArXiv}, abs/2506.21539, 2025.

\bibitem[Zhang et~al.(2025{\natexlab{c}})Zhang, Chen, Xu, Huang, Zhou, Yuan, Cai, Huang, Quan, Xu, and Zhang]{Zhang20254DVLASV}
Jiahui Zhang, Yurui Chen, Yueming Xu, Ze~Huang, Yanpeng Zhou, Yuan Yuan, Xinyue Cai, Guowei Huang, Xingyue Quan, Hang Xu, and Li~Zhang.
\newblock 4d-vla: Spatiotemporal vision-language-action pretraining with cross-scene calibration.
\newblock \emph{ArXiv}, abs/2506.22242, 2025{\natexlab{c}}.

\bibitem[Sendai et~al.(2025)Sendai, Alvarez, Matsushima, Matsuo, and Iwasawa]{Sendai2025LeaveNO}
Kohei Sendai, Maxime Alvarez, Tatsuya Matsushima, Yutaka Matsuo, and Yusuke Iwasawa.
\newblock Leave no observation behind: Real-time correction for vla action chunks.
\newblock \emph{ArXiv}, abs/2509.23224, 2025.

\bibitem[Jiang et~al.(2026)Jiang, Zhou, Jiang, Huang, Wei, Chen, Zhou, Guo, Lin, Zhang, Wang, Li, Yu, and Zhao]{Jiang2026WoVRWM}
Zhennan Jiang, Shan Zhou, Yutong Jiang, Zefang Huang, Mingjie Wei, Yuhui Chen, Tianxing Zhou, Zhen Guo, Hao Lin, Quanlu Zhang, Yu~Wang, Haoran Li, Chao Yu, and Dongbin Zhao.
\newblock Wovr: World models as reliable simulators for post-training vla policies with rl.
\newblock \emph{ArXiv}, abs/2602.13977, 2026.

\bibitem[Zhi et~al.(2025)Zhi, Chen, Zhou, Yu, Wu, Han, and Tan]{Zhi20253DFlowActionLC}
Hongyan Zhi, Peihao Chen, Siyuan Zhou, Dongjie Yu, Quanxi Wu, Lei Han, and Mingkui Tan.
\newblock 3dflowaction: Learning cross-embodiment manipulation from 3d flow world model.
\newblock \emph{ArXiv}, abs/2506.06199, 2025.

\bibitem[Kambara et~al.(2026)Kambara, Seno, Kaichi, Wang, and Sugiura]{Kambara2026LILACLO}
Motonari Kambara, Koki Seno, Tomoya Kaichi, Yanan Wang, and Komei Sugiura.
\newblock Lilac: Language-conditioned object-centric optical flow for open-loop trajectory generation.
\newblock \emph{IEEE Robotics and Automation Letters}, 11:\penalty0 6767--6774, 2026.

\bibitem[Song et~al.(2025)Song, Chen, Ding, Zhao, Zhao, Zhong, Ge, Ma, and Li]{Song2025PDVLAAV}
Wenxuan Song, Jiayi Chen, Pengxiang Ding, Han Zhao, Wei Zhao, Zhide Zhong, Zongyuan Ge, Jun Ma, and Haoang Li.
\newblock Pd-vla: Accelerating vision-language-action model integrated with action chunking via parallel decoding.
\newblock \emph{2025 IEEE/RSJ International Conference on Intelligent Robots and Systems (IROS)}, pages 13162--13169, 2025.

\bibitem[Liu et~al.(2025{\natexlab{a}})Liu, Zhang, Dong, Ye, Yuan, Yu, Yin, Lu, Shi, Yu, Zheng, Jiang, Gong, Qiu, and Zhao]{Liu2025FASTerTE}
Yicheng Liu, Shiduo Zhang, Zibin Dong, Baijun Ye, Tianyuan Yuan, Xiaopeng Yu, Linqi Yin, Chenhao Lu, Junhao Shi, Luca Jiang-Tao Yu, Liangtao Zheng, Tao Jiang, Jingjing Gong, Xipeng Qiu, and Hang Zhao.
\newblock Faster: Toward efficient autoregressive vision language action modeling via neural action tokenization.
\newblock \emph{ArXiv}, abs/2512.04952, 2025{\natexlab{a}}.

\bibitem[Black et~al.(2025{\natexlab{a}})Black, Galliker, and Levine]{Black2025RealTimeEO}
Kevin Black, Manuel~Y. Galliker, and Sergey Levine.
\newblock Real-time execution of action chunking flow policies.
\newblock \emph{ArXiv}, abs/2506.07339, 2025{\natexlab{a}}.

\bibitem[Liu et~al.(2024{\natexlab{a}})Liu, Hamid, Xie, Lee, Du, and Finn]{Liu2024BidirectionalDI}
Yuejiang Liu, Jubayer~Ibn Hamid, Annie Xie, Yoonho Lee, Maximilian Du, and Chelsea Finn.
\newblock Bidirectional decoding: Improving action chunking via guided test-time sampling.
\newblock In \emph{International Conference on Learning Representations}, 2024{\natexlab{a}}.

\bibitem[Wen et~al.(2026)Wen, Zhu, Zhang, Xia, Gao, and Li]{Wen2026AdaptiveAC}
Qingpeng Wen, Haomin Zhu, Yuepeng Zhang, Linzhong Xia, Bo~Gao, and Zhuozhen Li.
\newblock Adaptive action chunking for robotic imitation learning.
\newblock \emph{Biomimetics}, 2026.

\bibitem[Huang et~al.(2026)Huang, Zhang, Liu, Song, Tang, and Ma]{Huang2026TICVLAAT}
Zhiyu Huang, Yun Zhang, Johnson Liu, Rui Song, Chen Tang, and Jiaqi Ma.
\newblock Tic-vla: A think-in-control vision-language-action model for robot navigation in dynamic environments.
\newblock \emph{ArXiv}, abs/2602.02459, 2026.

\bibitem[Brohan et~al.(2022)Brohan, Brown, Carbajal, Chebotar, Dabis, Finn, Gopalakrishnan, Hausman, Herzog, Hsu, Ibarz, Ichter, Irpan, Jackson, Jesmonth, Joshi, Julian, Kalashnikov, Kuang, Leal, Lee, Levine, Lu, Malla, Manjunath, Mordatch, Nachum, Parada, Peralta, Perez, Pertsch, Quiambao, Rao, Ryoo, Salazar, Sanketi, Sayed, Singh, Sontakke, Stone, Tan, Tran, Vanhoucke, Vega, Vuong, Xia, Xiao, Xu, Xu, Yu, and Zitkovich]{Brohan2022RT1RT}
Anthony Brohan, Noah Brown, Justice Carbajal, Yevgen Chebotar, Joseph Dabis, Chelsea Finn, Keerthana Gopalakrishnan, Karol Hausman, Alexander Herzog, Jasmine Hsu, Julian Ibarz, Brian Ichter, Alex Irpan, Tomas Jackson, Sally Jesmonth, Nikhil~J. Joshi, Ryan~C. Julian, Dmitry Kalashnikov, Yuheng Kuang, Isabel Leal, Kuang-Huei Lee, Sergey Levine, Yao Lu, Utsav Malla, Deeksha Manjunath, Igor Mordatch, Ofir Nachum, Carolina Parada, Jodilyn Peralta, Emily Perez, Karl Pertsch, Jornell Quiambao, Kanishka Rao, Michael~S. Ryoo, Grecia Salazar, Pannag~R. Sanketi, Kevin Sayed, Jaspiar Singh, Sumedh~Anand Sontakke, Austin Stone, Clayton Tan, Huong Tran, Vincent Vanhoucke, Steve Vega, Quan~Ho Vuong, F.~Xia, Ted Xiao, Peng Xu, Sichun Xu, Tianhe Yu, and Brianna Zitkovich.
\newblock Rt-1: Robotics transformer for real-world control at scale.
\newblock \emph{ArXiv}, abs/2212.06817, 2022.

\bibitem[Padalkar et~al.(2023)Padalkar, Pooley, Jain, Bewley, Herzog, Irpan, Khazatsky, Rai, Singh, Brohan, Raffin, Wahid, Burgess-Limerick, Kim, Sch{\"o}lkopf, Ichter, Lu, Xu, Finn, Xu, Chi, Huang, Chan, Pan, Fu, Devin, Driess, Pathak, Shah, B{\"u}chler, Kalashnikov, Sadigh, Johns, Ceola, Xia, Stulp, Zhou, Sukhatme, Salhotra, Yan, Schiavi, Su, Fang, Shi, Amor, Christensen, Furuta, Walke, Fang, Mordatch, Radosavovic, Leal, Liang, Kim, Schneider, Hsu, Bohg, Bingham, Wu, Wu, Sun, Luo, Gu, Tan, Oh, Malik, Tompson, Yang, Lim, Silv{\'e}rio, Han, Rao, Pertsch, Hausman, Go, Gopalakrishnan, Goldberg, Byrne, Oslund, Kawaharazuka, Zhang, Majd, Rana, Srinivasan, Chen, Pinto, Tan, Ott, Lee, Tomizuka, Du, Ahn, Zhang, Ding, Srirama, Sharma, Kim, Irshad, Kanazawa, Hansen, Heess, Joshi, Suenderhauf, Palo, Shafiullah, Mees, Kroemer, Sanketi, Wohlhart, Xu, Sermanet, Sundaresan, Vuong, Rafailov, Tian, Doshi, Mendonca, Shah, Hoque, Julian, Bustamante, Kirmani, Levine, Moore, Bahl, Dass, Song, Xu, Haldar, Adebola, Guist,
  Nasiriany, Schaal, Welker, Tian, Dasari, Belkhale, Osa, Harada, Matsushima, Xiao, Yu, Ding, Davchev, Zhao, Armstrong, Darrell, Jain, Vanhoucke, Zhan, Zhou, Burgard, Chen, Wang, Zhu, Li, Lu, Chebotar, Zhou, Zhu, Xu, Wang, Bisk, Cho, Lee, Cui, Wu, Tang, Zhu, Li, Iwasawa, Matsuo, Xu, and Cui]{Padalkar2023OpenXR}
Abhishek Padalkar, Acorn Pooley, Ajinkya Jain, Alex Bewley, Alex Herzog, Alex Irpan, Alexander Khazatsky, Anant Rai, Anikait Singh, Anthony Brohan, Antonin Raffin, Ayzaan Wahid, Ben Burgess-Limerick, Beomjoon Kim, Bernhard Sch{\"o}lkopf, Brian Ichter, Cewu Lu, Charles Xu, Chelsea Finn, Chenfeng Xu, Cheng Chi, Chenguang Huang, Christine Chan, Chuer Pan, Chuyuan Fu, Coline Devin, Danny Driess, Deepak Pathak, Dhruv Shah, Dieter B{\"u}chler, Dmitry Kalashnikov, Dorsa Sadigh, Edward Johns, Federico Ceola, Fei Xia, Freek Stulp, Gaoyue Zhou, Gaurav~S. Sukhatme, Gautam Salhotra, Ge~Yan, Giulio Schiavi, Hao Su, Haoshu Fang, Haochen Shi, Heni~Ben Amor, Henrik~I Christensen, Hiroki Furuta, Homer~Rich Walke, Hongjie Fang, Igor Mordatch, Ilija Radosavovic, Isabel Leal, Jacky Liang, Jaehyung Kim, Jan Schneider, Jasmine Hsu, Jeannette Bohg, Jeff Bingham, Jiajun Wu, Jialin Wu, Jiankai Sun, Jianlan Luo, Jiayuan Gu, Jie Tan, Jihoon Oh, Jitendra Malik, Jonathan Tompson, Jonathan Yang, Joseph~J. Lim, Jo{\~a}o Silv{\'e}rio,
  Junhyek Han, Kanishka Rao, Karl Pertsch, Karol Hausman, Keegan Go, Keerthana Gopalakrishnan, Ken Goldberg, Kendra Byrne, Kenneth Oslund, Kento Kawaharazuka, Kevin Zhang, Keyvan Majd, Krishan Rana, Krishna~Parasuram Srinivasan, Lawrence~Yunliang Chen, Lerrel Pinto, Liam Tan, Lionel Ott, Lisa Lee, Masayoshi Tomizuka, Maximilian Du, Michael Ahn, Mingtong Zhang, Mingyu Ding, Mohan~Kumar Srirama, Mohit Sharma, Moo~Jin Kim, Muhammad~Zubair Irshad, Naoaki Kanazawa, Nicklas Hansen, Nicolas Manfred~Otto Heess, Nikhil~J. Joshi, Niko Suenderhauf, Norman~Di Palo, Nur~Muhammad Shafiullah, Oier Mees, Oliver Kroemer, Pannag~R. Sanketi, Paul Wohlhart, Peng Xu, Pierre Sermanet, Priya Sundaresan, Quan~Ho Vuong, Rafael Rafailov, Ran Tian, Ria Doshi, Russell Mendonca, Rutav Shah, Ryan Hoque, Ryan~C. Julian, Samuel Bustamante, Sean Kirmani, Sergey Levine, Sherry Moore, Shikhar Bahl, Shivin Dass, Shuran Song, Sichun Xu, Siddhant Haldar, Simeon Adebola, Simon Guist, Soroush Nasiriany, Stefan Schaal, Stefan Welker, Stephen Tian,
  Sudeep Dasari, Suneel Belkhale, Takayuki Osa, Tatsuya Harada, Tatsuya Matsushima, Ted Xiao, Tianhe Yu, Tianli Ding, Todor Davchev, Tony Zhao, Travis Armstrong, Trevor Darrell, Vidhi Jain, Vincent Vanhoucke, Wei Zhan, Wenxuan Zhou, Wolfram Burgard, Xi~Chen, Xiaolong Wang, Xinghao Zhu, Xuanlin Li, Yao Lu, Yevgen Chebotar, Yifan Zhou, Yifeng Zhu, Ying Xu, Yixuan Wang, Yonatan Bisk, Yoonyoung Cho, Youngwoon Lee, Yuchen Cui, Yueh-Hua Wu, Yujin Tang, Yuke Zhu, Yunzhu Li, Yusuke Iwasawa, Yutaka Matsuo, Zhuo Xu, and Zichen~Jeff Cui.
\newblock Open x-embodiment: Robotic learning datasets and rt-x models : Open x-embodiment collaboration0.
\newblock \emph{2024 IEEE International Conference on Robotics and Automation (ICRA)}, pages 6892--6903, 2023.

\bibitem[Intelligence et~al.(2025)Intelligence, Black, Brown, Darpinian, Dhabalia, Driess, Esmail, Equi, Finn, Fusai, Galliker, Ghosh, Groom, Hausman, Ichter, Jakubczak, Jones, Ke, LeBlanc, Levine, Li-Bell, Mothukuri, Nair, Pertsch, Ren, Shi, Smith, Springenberg, Stachowicz, Tanner, Vuong, Walke, Walling, Wang, Yu, and Zhilinsky]{Intelligence202505AV}
Physical Intelligence, Kevin Black, Noah Brown, James Darpinian, Karan Dhabalia, Danny Driess, Adnan Esmail, Michael Equi, Chelsea Finn, Niccolo Fusai, Manuel~Y. Galliker, Dibya Ghosh, Lachy Groom, Karol Hausman, Brian Ichter, Szymon Jakubczak, Tim Jones, Liyiming Ke, Devin LeBlanc, Sergey Levine, Adrian Li-Bell, Mohith Mothukuri, Suraj Nair, Karl Pertsch, Allen~Z. Ren, Lucy~Xiaoyang Shi, Laura Smith, Jost~Tobias Springenberg, Kyle Stachowicz, James Tanner, Quan Vuong, Homer~Rich Walke, Anna Walling, Haohuan Wang, Lili Yu, and Ury Zhilinsky.
\newblock $\pi$0.5: a vision-language-action model with open-world generalization.
\newblock \emph{ArXiv}, abs/2504.16054, 2025.

\bibitem[Shukor et~al.(2025)Shukor, Aubakirova, Capuano, Kooijmans, Palma, Zouitine, Aractingi, Pascal, Russi, Marafioti, Alibert, Cord, Wolf, and Cad{\`e}ne]{Shukor2025SmolVLAAV}
Mustafa Shukor, Dana Aubakirova, Francesco Capuano, Pepijn Kooijmans, Steven Palma, Adil Zouitine, Michel Aractingi, Caroline Pascal, Martino Russi, Andr{\'e}s Marafioti, Simon Alibert, Matthieu Cord, Thomas Wolf, and R{\'e}mi Cad{\`e}ne.
\newblock Smolvla: A vision-language-action model for affordable and efficient robotics.
\newblock \emph{ArXiv}, abs/2506.01844, 2025.

\bibitem[Liu et~al.(2025{\natexlab{b}})Liu, Chen, An, Liu, Zhang, Gu, Li, Guo, Chen, Liu, Hou, Zhao, alex Zhou, Heng, and Zhang]{Liu2025HybridVLACD}
Jiaming Liu, Hao Chen, Pengju An, Zhuoyang Liu, Renrui Zhang, Chenyang Gu, Xiaoqi Li, Ziyu Guo, Sixiang Chen, Mengzhen Liu, Chengkai Hou, Mengdi Zhao, KC~alex Zhou, Pheng-Ann Heng, and Shanghang Zhang.
\newblock Hybridvla: Collaborative diffusion and autoregression in a unified vision-language-action model.
\newblock \emph{ArXiv}, abs/2503.10631, 2025{\natexlab{b}}.

\bibitem[Liu et~al.(2024{\natexlab{b}})Liu, Wu, Li, Tan, Chen, Wang, Xu, Su, and Zhu]{Liu2024RDT1BAD}
Songming Liu, Lingxuan Wu, Bangguo Li, Hengkai Tan, Huayu Chen, Zhengyi Wang, Ke~Xu, Hang Su, and Jun Zhu.
\newblock Rdt-1b: a diffusion foundation model for bimanual manipulation.
\newblock \emph{ArXiv}, abs/2410.07864, 2024{\natexlab{b}}.

\bibitem[Zhang et~al.(2025{\natexlab{d}})Zhang, Pfrommer, Pan, Matni, and Simchowitz]{Zhang2025ActionCA}
Thomas~T. Zhang, Daniel Pfrommer, Chaoyi Pan, Nikolai Matni, and Max Simchowitz.
\newblock Action chunking and exploratory data collection yield exponential improvements in behavior cloning for continuous control.
\newblock 2025{\natexlab{d}}.
\newblock URL \url{https://api.semanticscholar.org/CorpusID:280254015}.

\bibitem[Mu et~al.(2021)Mu, Ling, Xiang, Yang, Li, Tao, Huang, Jia, and Su]{Mu2021ManiSkillGM}
Tongzhou Mu, Z.~Ling, Fanbo Xiang, Derek Yang, Xuanlin Li, Stone Tao, Zhiao Huang, Zhiwei Jia, and Hao Su.
\newblock Maniskill: Generalizable manipulation skill benchmark with large-scale demonstrations.
\newblock In \emph{NeurIPS Datasets and Benchmarks}, 2021.

\bibitem[Gu et~al.(2023)Gu, Xiang, Li, Ling, Liu, Mu, Tang, Tao, Wei, Yao, Yuan, Xie, Huang, Chen, and Su]{Gu2023ManiSkill2AU}
Jiayuan Gu, Fanbo Xiang, Xuanlin Li, Z.~Ling, Xiqiang Liu, Tongzhou Mu, Yihe Tang, Stone Tao, Xinyue Wei, Yuan Yao, Xiao Yuan, Pengwei Xie, Zhiao Huang, Rui Chen, and Hao Su.
\newblock Maniskill2: A unified benchmark for generalizable manipulation skills.
\newblock \emph{ArXiv}, abs/2302.04659, 2023.

\bibitem[Nasiriany et~al.(2024)Nasiriany, Maddukuri, Zhang, Parikh, Lo, Joshi, Mandlekar, and Zhu]{Nasiriany2024RoboCasaLS}
Soroush Nasiriany, Abhiram Maddukuri, Lance Zhang, Adeet Parikh, Aaron Lo, Abhishek Joshi, Ajay Mandlekar, and Yuke Zhu.
\newblock Robocasa: Large-scale simulation of everyday tasks for generalist robots.
\newblock \emph{ArXiv}, abs/2406.02523, 2024.

\bibitem[Zhang et~al.(2024{\natexlab{a}})Zhang, Xu, Liu, Yu, Li, Gao, Fei, Yin, Wu, Jiang, and Qiu]{Zhang2024VLABenchAL}
Shiduo Zhang, Zhe Xu, Peiju Liu, Xiaopeng Yu, Yuan Li, Qinghui Gao, Zhaoye Fei, Zhangyue Yin, Zuxuan Wu, Yu-Gang Jiang, and Xipeng Qiu.
\newblock Vlabench: A large-scale benchmark for language-conditioned robotics manipulation with long-horizon reasoning tasks.
\newblock \emph{2025 IEEE/CVF International Conference on Computer Vision (ICCV)}, pages 11142--11152, 2024{\natexlab{a}}.

\bibitem[Burgess-Limerick et~al.(2022)Burgess-Limerick, Lehnert, Leitner, and Corke]{BurgessLimerick2022DGBenchAO}
Ben Burgess-Limerick, Christopher~F. Lehnert, J.~Leitner, and Peter Corke.
\newblock Dgbench: An open-source, reproducible benchmark for dynamic grasping.
\newblock \emph{2022 IEEE/RSJ International Conference on Intelligent Robots and Systems (IROS)}, pages 3218--3224, 2022.

\bibitem[Hassan et~al.(2024)Hassan, Stapf, Rahimi, Rezende, Haghighi, Br{\"u}ggemann, Katircioglu, Zhang, Chen, Saha, Cannici, Aljalbout, Ye, Wang, Davtyan, Salzmann, Scaramuzza, Pollefeys, Favaro, and Alahi]{Hassan2024GEMAG}
Mariam Hassan, Sebastian Stapf, Ahmad Rahimi, Pedro Martelleto~Bressane Rezende, Yasaman Haghighi, David Br{\"u}ggemann, Isinsu Katircioglu, Lin Zhang, Xiaoran Chen, Suman Saha, Marco Cannici, Elie Aljalbout, Botao Ye, Xi~Wang, Aram Davtyan, Mathieu Salzmann, Davide Scaramuzza, Marc Pollefeys, Paolo Favaro, and Alexandre Alahi.
\newblock Gem: A generalizable ego-vision multimodal world model for fine-grained ego-motion, object dynamics, and scene composition control.
\newblock \emph{2025 IEEE/CVF Conference on Computer Vision and Pattern Recognition (CVPR)}, pages 22404--22415, 2024.

\bibitem[Wang et~al.(2026)Wang, Wang, Xu, Ye, Zhang, Tian, Zhan, Zhu, Lu, and Yang]{Wang2026LaMPLV}
Xinkai Wang, Chenyi Wang, Yifu Xu, Ming Ye, Fugang Zhang, Jialin Tian, Xinyu Zhan, Lifeng Zhu, Cewu Lu, and Lixin Yang.
\newblock Lamp: Learning vision-language-action policies with 3d scene flow as latent motion prior.
\newblock 2026.

\bibitem[Fan et~al.(2026)Fan, Liu, Li, Ren, Li, Zhang, Ding, and Deng]{Fan2026FUTUREVLAFU}
Jingjing Fan, Yushan Liu, Shoujie Li, Botao Ren, Siyuan Li, Xiao-Ping Zhang, Wenbo Ding, and Zhidong Deng.
\newblock Future-vla: Forecasting unified trajectories under real-time execution.
\newblock \emph{ArXiv}, abs/2602.15882, 2026.

\bibitem[Liu et~al.(2026{\natexlab{a}})Liu, Tan, Zhu, Li, Li, Yang, and Shen]{Liu2026SelfCorrectingVO}
Chen-Yu Liu, Wentao Tan, Lei Zhu, Fengling Li, Jingjing Li, Guoli Yang, and Heng~Tao Shen.
\newblock Self-correcting vla: Online action refinement via sparse world imagination.
\newblock \emph{ArXiv}, abs/2602.21633, 2026{\natexlab{a}}.

\bibitem[Xu et~al.(2025)Xu, Wang, Xia, Zhu, Huang, and Xu]{Xu2025VLACacheEV}
Siyu Xu, Yunke Wang, Chenghao Xia, Di~Zhu, Tao Huang, and Chang Xu.
\newblock Vla-cache: Efficient vision-language-action manipulation via adaptive token caching.
\newblock 2025.

\bibitem[Tan et~al.(2025)Tan, Yang, Ye, Zheng, Bai, Wang, Hao, and Chen]{Tan2025ThinkTA}
Xudong Tan, Yaoxin Yang, Peng Ye, Jiali Zheng, Bizhe Bai, Xinyi Wang, Jia Hao, and Tao Chen.
\newblock Think twice, act once: Token-aware compression and action reuse for efficient inference in vision-language-action models.
\newblock \emph{ArXiv}, abs/2505.21200, 2025.

\bibitem[Liang et~al.(2025)Liang, Li, Yang, Wu, Mao, Nian, Pei, Zhou, Yang, Pang, et~al.]{liang2025discrete}
Zhixuan Liang, Yizhuo Li, Tianshuo Yang, Chengyue Wu, Sitong Mao, Tian Nian, Liuao Pei, Shunbo Zhou, Xiaokang Yang, Jiangmiao Pang, et~al.
\newblock Discrete diffusion vla: Bringing discrete diffusion to action decoding in vision-language-action policies.
\newblock \emph{arXiv preprint arXiv:2508.20072}, 2025.

\bibitem[Zhang et~al.(2024{\natexlab{b}})Zhang, Guo, Chen, Wang, Hu, Shi, and Chen]{zhang2024hirt}
Jianke Zhang, Yanjiang Guo, Xiaoyu Chen, Yen-Jen Wang, Yucheng Hu, Chengming Shi, and Jianyu Chen.
\newblock Hirt: Enhancing robotic control with hierarchical robot transformers.
\newblock \emph{arXiv preprint arXiv:2410.05273}, 2024{\natexlab{b}}.

\bibitem[Black et~al.(2025{\natexlab{b}})Black, Ren, Equi, and Levine]{Black2025TrainingTimeAC}
Kevin Black, Allen~Z. Ren, Michael Equi, and Sergey Levine.
\newblock Training-time action conditioning for efficient real-time chunking.
\newblock \emph{ArXiv}, abs/2512.05964, 2025{\natexlab{b}}.

\bibitem[Liu et~al.(2026{\natexlab{b}})Liu, Yu, Zhao, Li, Zhang, Li, Wu, Hu, Xie, Guo, Wang, and Gao]{Liu2026LearningNC}
Yufeng Liu, Hang Yu, Juntu Zhao, Bocheng Li, Di~Zhang, Ming-Zhe Li, Wenxuan Wu, Yingdong Hu, Junyuan Xie, Junliang Guo, Dequan Wang, and Yang Gao.
\newblock Learning native continuation for action chunking flow policies.
\newblock \emph{ArXiv}, abs/2602.12978, 2026{\natexlab{b}}.

\end{thebibliography}


\appendix

\section{Full Closed-Form Mathematical Derivation}\label{app:derivation}

This appendix provides the complete mathematical derivation of the Pace-and-Path Correction operator summarized in Sections~\ref{method}. All results are derived from the single quadratic cost introduced in Section~\ref{sec:problem}; no additional assumptions beyond A1--A3 (stated below) are introduced.

\subsection{Assumptions}

Three working assumptions underlie the derivation:
\begin{enumerate}
    \item[\textbf{A1.}] \textit{Quasi-stationary plan.} The per-step delta $\Delta p_i \approx \Delta p$ is approximately constant within the executed chunk window $i = 0, \ldots, K-1$.
    \item[\textbf{A2.}] \textit{Slowly-varying disturbance.} The velocity $v$ and direction $\hat{d}$ are approximately constant over the $K$ executed env-steps (relaxed to affine variation in Section~\ref{app:second_order}).
    \item[\textbf{A3.}] \textit{Small rotation.} Per-step rotations are small enough that xyz deltas are additive across env-steps.
\end{enumerate}
All three degrade gracefully: violations reduce the accuracy of the optimum but do not destabilize the operator, since the chunk boundary refresh resets all signals.

\subsection{Cost Function}\label{app:cost}

Under A1, the corrected delta at env-step $k$ is $u_k = \alpha \Delta p + \delta_k$ with $\alpha \ge 1$ and $\delta_k \in \mathbb{R}^3$. The cumulative arm position is $p_j = \sum_{k=0}^{j-1} u_k = j\alpha\Delta p + \sigma_j$ where $\sigma_j := \sum_{k=0}^{j-1}\delta_k$. The ideal tracking trajectory under disturbance $(v, \hat{d})$ is $\tilde{p}_j = j(\Delta p + v\hat{d})$. The cost balances per-waypoint tracking against per-step offset effort:
\begin{equation}\label{eq:cost}
    L(\alpha, \{\delta_k\}) = \frac{1}{2}\sum_{j=1}^{K}\|p_j - \tilde{p}_j\|^2 + \frac{1}{2}\sum_{k=0}^{K-1}\|\delta_k\|^2.
\end{equation}

The penalty is on $\delta_k$ only, not on $\alpha$, because $\alpha$ moves the arm along the planned direction (temporal compression with no directional deviation), whereas $\delta_k$ introduces off-plan spatial offset. This asymmetry is the unique choice satisfying three invariants simultaneously: (i) uniqueness of the optimum, (ii) degeneracy to the baseline VLA at $v=0$, and (iii) the cosine projection structure in $\alpha^\star$.

\subsection{Reduced Quantities and Stationarity Conditions}\label{app:stationarity}

Define the residual disturbance $A := v\hat{d} - (\alpha - 1)\Delta p$. The tracking error simplifies to
\begin{equation}
    e_j := p_j - \tilde{p}_j = -jA + \sigma_j, \qquad j = 1, \ldots, K.
\end{equation}

Setting $\partial L / \partial \alpha = 0$ and $\partial L / \partial \delta_k = 0$ yields the joint stationarity conditions:
\begin{align}
    \frac{\partial L}{\partial \alpha} &= -\Delta p \cdot \sum_{j=1}^{K} j\,e_j = 0, \label{eq:stat_alpha}\\[4pt]
    \frac{\partial L}{\partial \delta_k} &= \delta_k + \sum_{j=k+1}^{K} e_j = 0, \qquad k = 0, \ldots, K-1. \label{eq:stat_delta}
\end{align}

\subsection{Derivation of $\alpha^\star$ (Pace Channel)}\label{app:alpha}

The cost~\eqref{eq:cost} is rotationally invariant in $\mathbb{R}^3$. At the optimum, every $\delta_k$ inherits the direction of $A$: $\delta_k = c_k A$ for some scalar $c_k$. Consequently $\sigma_j$ and $e_j$ are parallel to $A$, so $\sum_j j\,e_j$ is also parallel to $A$. Condition~\eqref{eq:stat_alpha} then collapses to
\begin{equation}
    \Delta p \cdot A = 0.
\end{equation}

Expanding $A = v\hat{d} - (\alpha - 1)\Delta p$:
\begin{equation}
    \Delta p \cdot \big(v\hat{d} - (\alpha - 1)\Delta p\big) = 0 \quad\Longrightarrow\quad (\alpha - 1)\|\Delta p\|^2 = v\,(\hat{d} \cdot \Delta p) = v\|\Delta p\|\cos\theta,
\end{equation}
where $\cos\theta := \hat{d} \cdot \widehat{\Delta p}$. Solving:
\begin{equation}\label{eq:alpha_star}
    \boxed{\;\alpha^\star = 1 + \frac{v\cos\theta}{\|\Delta p\|}.\;}
\end{equation}

Substituting $\alpha^\star$ back into $A$ yields the orthogonal residual:
\begin{equation}\label{eq:A_star}
    A^\star = v\hat{d} - v\cos\theta\,\widehat{\Delta p} = v\hat{d}_\perp,
\end{equation}
which lies entirely perpendicular to the planned direction $\widehat{\Delta p}$.

\textit{Clamping.} The physical constraint $\alpha \ge 1$ is violated when $\cos\theta < 0$ (antagonistic motion). In this case $\alpha^\star$ is clamped to 1, and the full disturbance $v\hat{d}$ passes to the path channel. When $\alpha^\star > T/K$ (exceeds the chunk budget), the dynamic execution horizon $K_{\mathrm{exec}}$ absorbs the overflow (Section~\ref{app:kexec}).

\subsection{Derivation of $\delta_k^\star$ (Path Channel, Fibonacci Profile)}\label{app:fib}

The path channel handles $A^\star$, which $\alpha^\star$ cannot absorb. Differencing condition~\eqref{eq:stat_delta} in $k$ and using $\sigma_{j+1} - \sigma_j = \delta_j$ yields the 2D linear recurrence
\begin{equation}\label{eq:recurrence}
    \begin{pmatrix} \delta_{k+1} \\ e_{k+1} \end{pmatrix}
    = \underbrace{\begin{pmatrix} 2 & 1 \\ 1 & 1 \end{pmatrix}}_{M}
    \begin{pmatrix} \delta_k \\ e_k \end{pmatrix}
    - \begin{pmatrix} A^\star \\ A^\star \end{pmatrix},
\end{equation}
with boundary conditions $e_0 = 0$ and $\delta_K = 0$.

\paragraph{Eigenstructure.} The companion matrix $M$ has characteristic polynomial $\lambda^2 - 3\lambda + 1 = 0$, yielding eigenvalues
\begin{equation}
    \lambda_\pm = \frac{3 \pm \sqrt{5}}{2} = \varphi^{\pm 2}, \qquad \varphi = \frac{1+\sqrt{5}}{2} \;\text{(golden ratio)}.
\end{equation}

\paragraph{Particular solution.} Setting $\delta_{k+1} = \delta_k =: \delta^p$ and $e_{k+1} = e_k =: e^p$ in~\eqref{eq:recurrence} gives $(\delta^p, e^p) = (A^\star, 0)$.

\paragraph{Homogeneous solution.} The eigenvectors of $M$ are $w_+ = (\varphi, 1)^\top$ and $w_- = (1, -\varphi)^\top$. The general solution is
\begin{equation}
    \begin{pmatrix} \delta_k \\ e_k \end{pmatrix} = \begin{pmatrix} A^\star \\ 0 \end{pmatrix} + c_+ \varphi^{2k} \begin{pmatrix} \varphi \\ 1 \end{pmatrix} + c_- \varphi^{-2k} \begin{pmatrix} 1 \\ -\varphi \end{pmatrix}.
\end{equation}

\paragraph{Boundary conditions.} From $e_0 = 0$: $c_+ - \varphi\,c_- = 0$, so $c_+ = \varphi\,c_-$. From $\delta_K = 0$:
\begin{equation}
    A^\star + c_-\big(\varphi^{2K+1} + \varphi^{-2K}\big) = 0.
\end{equation}
Applying the identity $\varphi^n + (-\varphi)^{-n} = \sqrt{5}\,F_n$ for odd $n$ (where $F_n$ is the $n$-th Fibonacci number with $F_1 = F_2 = 1$), we obtain
\begin{equation}
    \varphi^{2K+1} + \varphi^{-(2K+1)} = \sqrt{5}\,F_{2K+1},
\end{equation}
noting that $\varphi^{-2K} = (-1)^{2K}\varphi^{-2K} = \varphi^{-2K}$ and adjusting signs gives $c_- = -A^\star / (\sqrt{5}\,F_{2K+1})$. Substituting back and collecting:
\begin{equation}\label{eq:delta_fib}
    \boxed{\;\delta_k^\star = \left(1 - \frac{F_{2k+1}}{F_{2K+1}}\right) A^\star, \qquad k = 0, \ldots, K-1.\;}
\end{equation}

\paragraph{Profile properties.} The Fibonacci ratio $F_{2k+1}/F_{2K+1}$ increases monotonically in $k$ from $F_1/F_{2K+1} = 1/F_{2K+1} \approx 0$ to $F_{2K-1}/F_{2K+1} \to \varphi^{-2} \approx 0.382$ as $K \to \infty$. Thus $\delta_0^\star \approx A^\star$ (near-full compensation at chunk start), decaying to $\delta_{K-1}^\star \approx 0.618\,A^\star$, with the terminal condition $\delta_K^\star = 0$ enforcing closure at the chunk boundary.

\paragraph{Verification.} For $K=2$: $F_5 = 5$, giving $\delta_0^\star = (1-1/5)A^\star = \tfrac{4}{5}A^\star$ and $\delta_1^\star = (1-3/5)A^\star = \tfrac{2}{5}A^\star$. Direct substitution into~\eqref{eq:stat_delta} confirms both conditions are satisfied.

\subsection{Orthogonal Decomposition}\label{app:ortho}

The two channels act on disjoint subspaces: $\alpha^\star$ absorbs $v\cos\theta\,\widehat{\Delta p}$ (the component of $v\hat{d}$ parallel to $\Delta p$), while $\delta_k^\star$ absorbs $A^\star = v\hat{d}_\perp$ (the perpendicular residual). The channels do not interact in each other's closed forms; both are fully determined by the chunk geometry $(\Delta p, K)$ and the dynamics signal $(v, \hat{d})$.

\subsection{General $\lambda$-Regularization}\label{app:lambda}

Replacing the unit weight on the effort term with a general $\lambda > 0$:
\begin{equation}
    L_\lambda = \frac{1}{2}\sum_{j=1}^{K}\|e_j\|^2 + \frac{\lambda}{2}\sum_{k=0}^{K-1}\|\delta_k\|^2.
\end{equation}

\paragraph{$\alpha^\star$ is $\lambda$-independent.} The factor $\lambda$ multiplies only the $\delta$-effort term. Since the rotation-invariance argument still forces $\delta_k \parallel A$ and hence $\Delta p \cdot A = 0$, the $\alpha^\star$ formula~\eqref{eq:alpha_star} holds unchanged for all $\lambda > 0$.

\paragraph{$\delta_k^\star(\lambda)$ in hyperbolic cosine form.} The modified recurrence becomes
\begin{equation}
    \begin{pmatrix} \delta_{k+1} \\ e_{k+1} \end{pmatrix}
    = \begin{pmatrix} 1+1/\lambda & 1/\lambda \\ 1 & 1 \end{pmatrix}
    \begin{pmatrix} \delta_k \\ e_k \end{pmatrix}
    + \begin{pmatrix} -A^\star/\lambda \\ -A^\star \end{pmatrix}.
\end{equation}
The eigenvalues $\mu_\pm$ satisfy $\mu + \mu^{-1} = 2 + 1/\lambda$, giving $\mu_+ = e^{\omega(\lambda)}$ with
\begin{equation}
    \omega(\lambda) = \mathrm{arccosh}\!\left(1 + \frac{1}{2\lambda}\right).
\end{equation}
Solving under the same boundary conditions:
\begin{equation}\label{eq:delta_cosh}
    \boxed{\;\delta_k^\star(\lambda) = \left(1 - \frac{\cosh\big((k+\tfrac{1}{2})\,\omega\big)}{\cosh\big((K+\tfrac{1}{2})\,\omega\big)}\right) A^\star.\;}
\end{equation}

\paragraph{Limiting cases.}
\begin{itemize}
    \item $\lambda = 1$: $\omega = 2\ln\varphi$, and $\cosh((2k{+}1)\ln\varphi) = \tfrac{\sqrt{5}}{2}F_{2k+1}$, recovering the Fibonacci profile~\eqref{eq:delta_fib}.
    \item $\lambda \to 0$: $\omega \to \infty$, so $\delta_k^\star \to A^\star$ for all $k < K$ (no penalty, full compensation).
    \item $\lambda \to \infty$: $\omega \to 0$, so $\delta_k^\star \to 0$ (high penalty, no spatial offset).
\end{itemize}

In practice, $\lambda$ is driven by the Bayesian confidence signal $K_{\mathrm{kal}}$ (Section~\ref{method}) via $\lambda = 1/K_{\mathrm{kal}}$, so uncertain observations ($K_{\mathrm{kal}} \to 0$) automatically suppress spatial offsets while leaving $\alpha^\star$ unaffected.

\subsection{Second-Order Extension (Acceleration)}\label{app:second_order}

We relax A2 to an affine-in-time disturbance $v(t) = v_0 + at$ with $t = 0, \ldots, K{-}1$. The cumulative target offset under midpoint integration becomes $jv_0 + \tfrac{1}{2}j^2 a$, modifying the ideal trajectory to $\tilde{p}_j = j\Delta p + (jv_0 + \tfrac{1}{2}j^2 a)\hat{d}$.

Define the two-component disturbance: $A := v_0\hat{d} - (\alpha-1)\Delta p$ (first-order) and $B := \tfrac{1}{2}a\hat{d}$ (second-order). The tracking error becomes $e_j = -jA - j^2 B + \sigma_j$.

\paragraph{$\alpha^\star$ under acceleration.} Setting $\partial L/\partial\alpha = 0$ and using the sums $S_2 = K(K{+}1)(2K{+}1)/6$ and $S_3 = [K(K{+}1)/2]^2$:
\begin{equation}\label{eq:alpha_second}
    \boxed{\;\alpha^\star = 1 + \frac{v_0\cos\theta_v}{\|\Delta p\|} + \frac{S_3}{2S_2}\cdot\frac{a\cos\theta_a}{\|\Delta p\|},\;}
\end{equation}
where $\cos\theta_v = \hat{d}_v \cdot \widehat{\Delta p}$ and $\cos\theta_a = \hat{d}_a \cdot \widehat{\Delta p}$. The coupling coefficient
\begin{equation}
    \frac{S_3}{2S_2} = \frac{3K(K+1)}{4(2K+1)} \xrightarrow{K \to \infty} \frac{3K}{8}
\end{equation}
scales linearly in $K$, reflecting the longer integration window over which acceleration accumulates. Setting $a = 0$ recovers~\eqref{eq:alpha_star}.

\paragraph{$\delta_k^\star$ under acceleration (Lucas profile).} The recurrence~\eqref{eq:recurrence} acquires an inhomogeneous term proportional to $B^\star := \tfrac{1}{2}a\hat{d}_{a,\perp}$ (the perpendicular component of the acceleration). Linearity of the recurrence yields an additive decomposition:
\begin{equation}\label{eq:delta_second}
    \boxed{\;\delta_k^\star = \underbrace{\left(1 - \frac{F_{2k+1}}{F_{2K+1}}\right)A^\star}_{\text{Fibonacci (first-order)}} + \underbrace{\Lambda_k(K)\,B^\star}_{\text{Lucas-polynomial (second-order)}},\;}
\end{equation}
where $\Lambda_k(K)$ is the Lucas-polynomial profile coefficient:
\begin{equation}
    \Lambda_k(K) = (2k+1) - L_{2k+1} + \frac{F_{2k+1}}{F_{2K+1}}\big(L_{2K+1} - (2K+1)\big),
\end{equation}
with $L_n$ denoting the Lucas numbers ($L_0 = 2$, $L_1 = 1$, $L_n = L_{n-1} + L_{n-2}$). The Lucas profile is the natural dual to Fibonacci on the same eigenvalue structure $\varphi^{\pm 2}$: the $j^2$-quadratic forcing activates the second homogeneous mode ($L_{2k+1}$ branch), which is silent in first-order because $L_0 = 2$ contradicts the boundary condition $e_0 = 0$.

\subsection{Dynamic Execution Horizon $K_{\mathrm{exec}}$}\label{app:kexec}

Rather than hard-clamping $\alpha^\star \le T/K$, we define a dynamic execution horizon that absorbs the overflow:
\begin{equation}
    K_{\mathrm{exec}}(\alpha) = \max\!\Big(K,\;\min\!\big(\lceil T/\alpha\rceil,\; T\big)\Big).
\end{equation}

At $\alpha = 1$, $K_{\mathrm{exec}} = T$ (full chunk consumed). As $\alpha$ increases, $K_{\mathrm{exec}}$ shrinks, reaching the floor $K$ at $\alpha \ge T/K$. The Fibonacci profile~\eqref{eq:delta_fib} then uses $K_{\mathrm{exec}}$ in place of $K$, adjusting the profile normalization to the actual execution window.

\subsection{Hierarchical 2-EMA Latch Stabilizer}\label{app:latch}

The closed-form correction is exact under A2. Irregular regimes (random walk, stop-and-go, teleport) violate A2 chronically, requiring a sustained cap on $K_{\mathrm{exec}}$ beyond what the single-chunk direction trust $\rho_t$ provides. The latch admits a single free hyperparameter $\beta_{\mathrm{in}}$; all other constants derive from the chunk geometry $(K, T)$.

\paragraph{Direction-shift trigger.} At each chunk reset $t$:
\begin{equation}
    \tau_t = \mathbf{1}\!\left[\rho_{\mathrm{gt}}(t) < \tfrac{1}{2}\right], \qquad \rho_{\mathrm{gt}}(t) = \max\!\left(0,\;\frac{v_t \cdot v_{t-1}}{\|v_t\|\,\|v_{t-1}\|}\right).
\end{equation}
The threshold $1/2$ is the natural midpoint of $\rho_{\mathrm{gt}} \in [0,1]$.

\paragraph{Outer EMA (chronic trigger rate).}
\begin{equation}
    C_t = \beta_{\mathrm{out}}\,\tau_t + (1 - \beta_{\mathrm{out}})\,C_{t-1}.
\end{equation}

\paragraph{Sticky factor.}
\begin{equation}
    s_t = \frac{C_t}{C_t + R_{\mathrm{TH}}}.
\end{equation}

\paragraph{Inner EMA with sticky-modulated decay.}
\begin{equation}
    L_t = \begin{cases}
        \beta_{\mathrm{in}} + (1-\beta_{\mathrm{in}})\,L_{t-1}, & \tau_t = 1,\\[4pt]
        \big[1 - \beta_{\mathrm{in}}(1-s_t)\big]\,L_{t-1}, & \tau_t = 0.
    \end{cases}
\end{equation}
Under chronic instability ($s_t \to 1$), the effective decay rate vanishes and $L_t$ holds near its current value. Under occasional triggers ($s_t \to 0$), $L_t$ decays at the standard rate $\beta_{\mathrm{in}}$.

\paragraph{Latch output.} $m_t = \mathbf{1}[L_t > L_{\mathrm{th}}]$. When $m_t = 1$, the execution horizon is capped: $K_{\mathrm{exec}} \le T/4$.

\paragraph{Derived constants.} All three internal thresholds derive from $\beta_{\mathrm{in}}$ and $(K, T)$:

\begin{enumerate}
    \item[(a)] \textit{Outer EMA rate.} Match the outer half-life to one chunk-budget cycle ($T/K$ chunks at $\alpha=1$). Solving $(1 - \beta_{\mathrm{out}})^{T/K} = 1/2$:
    \begin{equation}
        \beta_{\mathrm{out}} = 1 - 2^{-K/T}.
    \end{equation}
    For $K=2$, $T=16$: $\beta_{\mathrm{out}} \approx 0.083$.

    \item[(b)] \textit{Active threshold $L_{\mathrm{th}}$.} Under standard decay ($s_t = 0$), a single trigger at $t=0$ followed by $n$ non-trigger steps gives $L_n = \beta_{\mathrm{in}}(1-\beta_{\mathrm{in}})^n$. Setting $L_{\mathrm{th}} = L_2$ makes a single isolated trigger sustain the latch for exactly two chunks:
    \begin{equation}
        L_{\mathrm{th}} = \beta_{\mathrm{in}}(1-\beta_{\mathrm{in}})^2.
    \end{equation}
    For $\beta_{\mathrm{in}} = 0.3$: $L_{\mathrm{th}} \approx 0.147$.

    \item[(c)] \textit{Sticky reference.} $s_t$ reaches $1/2$ at $C_t = R_{\mathrm{TH}}$. The natural reference is the same scale as the active threshold:
    \begin{equation}
        R_{\mathrm{TH}} = L_{\mathrm{th}}.
    \end{equation}
\end{enumerate}

\paragraph{Regime behavior summary.}

\begin{center}
\small
\begin{tabular}{@{}llll@{}}
\toprule
Regime & $L_t$ behavior & $m_t$ & $K_{\mathrm{exec}}$ \\
\midrule
Stable (uniform/accel.) & $\to 0$ & 0 & full chunk budget \\
Single isolated event & spike, decays in $\sim$2 chunks & brief 1 & brief $T/4$ cap \\
Chronic (random walk) & sticks high ($s_t \to 1$) & persistent 1 & $T/4$ \\
Periodic (stop-and-go) & bursty with slow decay & intermittent 1 & intermittent cap \\
\bottomrule
\end{tabular}
\end{center}

\paragraph{Grasp reset.} When the TCP-to-object distance falls below the gripper half-span ($\|p_{\mathrm{tcp}} - p_{\mathrm{obj}}\| < r_{\mathrm{grip}}$), the object transitions from external dynamics to internal state of the manipulator. The latch state is reset: $(L_t, C_t) \leftarrow (0, 0)$.

\section{\textsc{MoveBench} Details}\label{app:movebench_details}

\begin{figure}[t]                                                         
    \centering                           
    \newcommand{\ycbcell}[2]{%
      \begin{subfigure}[t]{0.30\linewidth}                                  
        \centering\includegraphics[width=\linewidth]{ycb/#1.png}            
        \caption{#2}                                                        
      \end{subfigure}}                                                      
    \ycbcell{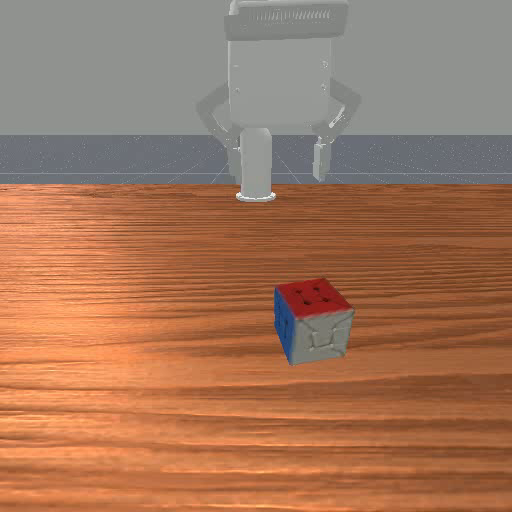}{rubik's cube}\hfill                           
    \ycbcell{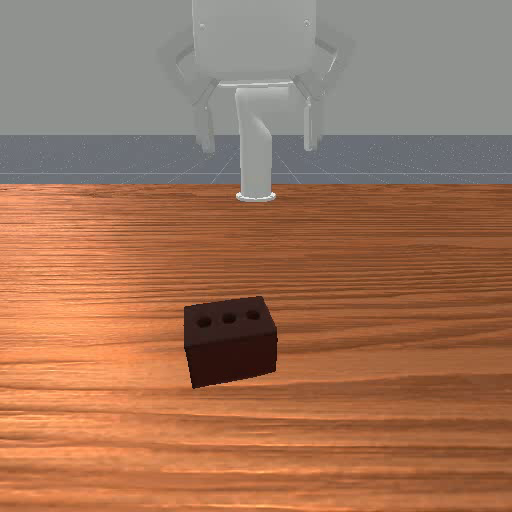}{foam brick}\hfill                              
    \ycbcell{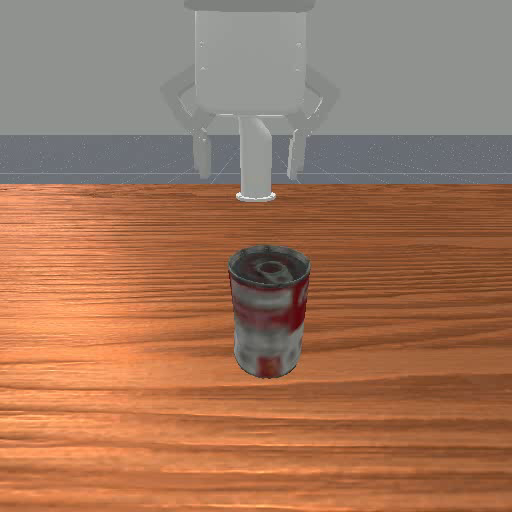}{tomato soup can}\\[6pt]                   
    \ycbcell{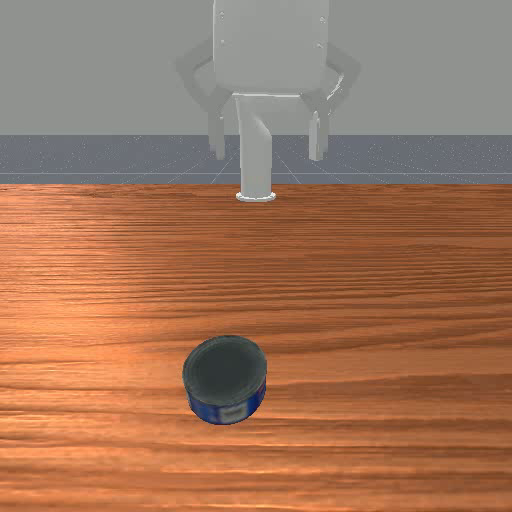}{tuna fish can}\hfill                        
    \ycbcell{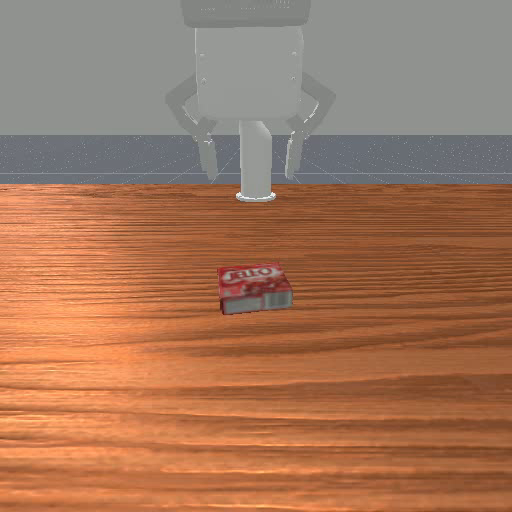}{gelatin box}\hfill                            
    \ycbcell{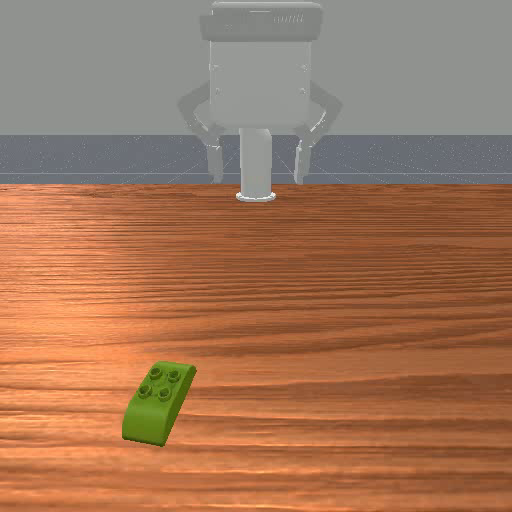}{lego brick}\\[6pt]                           
    \ycbcell{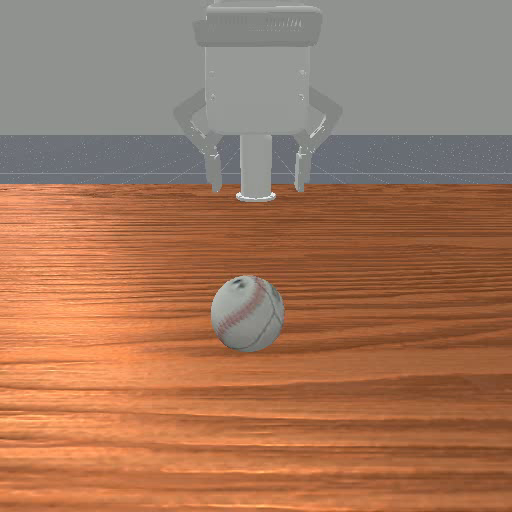}{baseball}\hfill                                
    \ycbcell{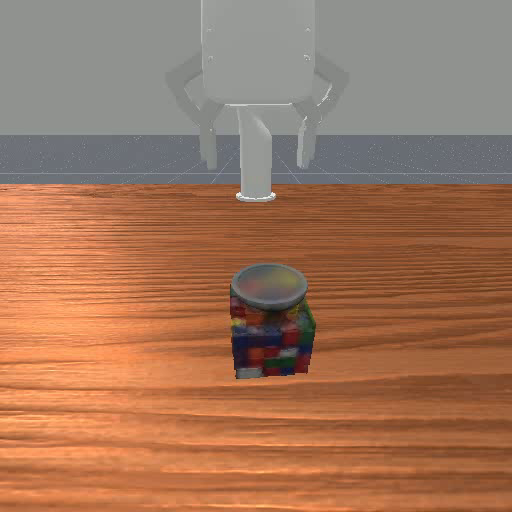}{jar}\hfill                          
    \ycbcell{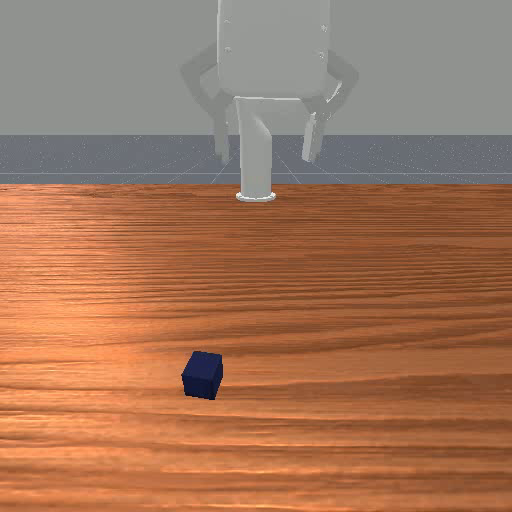}{wood block}                       
    \caption{The nine YCB objects sampled in \textsc{MoveBench}. Each panel 
  is the                                                                    
    base-camera frame at $t{=}0$ from a demonstration episode of the        
  corresponding                                                             
    task.}                                                                
    \label{fig:ycb_objects}                                                 
  \end{figure}  

This section expands the implementation details of \textsc{MoveBench} that are abbreviated in the main text. We control simulation at $20\,\text{Hz}$ via ManiSkill's SAPIEN backend, with $512{\times}512$ RGB streams from a fixed overhead and a wrist-mounted camera, $7$-DoF proprioception, and a language-instruction channel. Each episode admits at most $200$ environment steps ($10\,\text{s}$ wall-clock) and is judged successful when the target is grasped and lifted by $\geq 3\,\text{cm}$, matching the protocol established by the foundational VLA suites we benchmark against. The target is sampled uniformly from a pool of 9 YCB household objects covering a broad range of geometries (cube, brick, cylinder, sphere, flat box, jar) and physical sizes ($25\text{--}103\,\text{mm}$ along the dominant axis), exposing the  policy to grasp-pose ambiguity orthogonal to the motion variable. The sub-sections below detail the object pool, motion-regime parametrization, demonstration generation pipeline, and the diagnostic statistics we report.

\subsection{Object Pool}

The grasp target in every \textsc{MoveBench} environment is sampled uniformly
at episode reset from a fixed pool of nine YCB household objects, shown in
Fig.~\ref{fig:ycb_objects}. The pool is intentionally small but geometrically
heterogeneous, spanning the cube, rectangular brick, tall cylinder, flat
cylinder, flat box, elongated lego, sphere, jar, and tiny block primitives,
with object scales adjusted so that all targets fit within a $25$--$103\,\mathrm{mm}$
dominant-axis range and remain reachable by the xArm6 gripper. This diversity
ensures that the policy must commit to an object-specific grasp pose at the
moment of contact, which interacts non-trivially with the motion regime: a
motion handler that succeeds by approaching from a fixed direction collapses
once the pool forces approach-direction variation. The object identity is
exposed only through the natural-language instruction
``\texttt{Pick up the \{name\}.}'', so the policy must ground each name to its
visual appearance under whatever motion regime the episode draws.

\subsection{Motion Regimes}

Each environment is fully specified by a deterministic motion-update rule
that is sampled once per episode and applied to the target object every
simulation tick at $20\,\text{Hz}$. The seven dynamic environments instantiate
three families.

\textbf{Uniform translation.} The object is initialized with a single
uniformly-sampled direction in the table plane and a constant speed drawn
from a regime-specific range, $[1, 2]$, $[2, 4]$, and $[4, 8]\,\text{cm/s}$
for the easy, medium, and hard tier respectively. No further randomization is
applied during the episode, so the regime is fully characterized by the
initial speed magnitude.

\textbf{Accelerated motion.} The object is initialized with a low base speed
$v_0 \in [2, 3]\,\text{cm/s}$, common to all three tiers, and a per-episode
acceleration vector whose magnitude is drawn from $[2, 3]$, $[3, 5]$, and
$[5, 9]\,\text{cm/s}^2$ for easy, medium, and hard. Decoupling $v_0$ from the
acceleration tier ensures that any cross-tier performance gap is attributable
to the second-order signal alone, not to a confounded initial-speed shift.

\textbf{Irregular motion.} Three regime-change patterns probe non-stationary
behavior. \emph{Random Walk} maintains a constant $5\,\text{cm/s}$ speed but
resamples a fresh planar direction every $5$--$12$ ticks, simulating
continuous reactive disturbances at the timescale of a single chunk.
\emph{Stop-and-Go} alternates between $7\,\text{cm/s}$ uniform motion for
$3$--$7$ ticks and full pauses for $3$--$6$ ticks, presenting a binary
on/off velocity signal that rewards opportunistic grasping during pause
windows. \emph{Teleport} keeps the object stationary except for two
mid-episode discontinuities, the first scheduled within ticks $3$--$10$ and
each requiring a minimum displacement of $8\,\text{cm}$ from the current
position, an event whose magnitude exceeds any plausible per-chunk plan
correction. The three patterns are each provided at a single difficulty level
because they probe regime-change response rather than a continuously tunable
intensity scalar; introducing a difficulty axis would conflate the regime's
qualitative novelty with its quantitative magnitude.

\subsection{Episode and Workspace Configuration}
All environments share an identical scaffolding: a $7$-DoF xArm6 mounted at a
fixed table-side base pose, a planar tabletop with the target object
initialized within a square reachability region centered at the robot's
neutral grasp height, and two RGB cameras (a static overhead view and a
wrist-mounted view) streaming at $20\,\text{Hz}$. The observation passed to
the policy at each tick is the pair $(I_t^{\text{base}}, I_t^{\text{wrist}})$
together with a $7$-D proprioceptive state (end-effector position, axis-angle
orientation, gripper width). Every episode is capped at $200$ environment
steps ($10\,\text{s}$ wall-clock) and is judged successful only if the gripper
both contacts the target and lifts it by at least $3\,\text{cm}$ above its
resting height before the cap. Lift-only and approach-only events are
explicitly counted as failures, removing any pseudo-success that could arise
from a policy that brushes against the object without committing to a grasp.
Beyond the motion-regime parameters, episode-level randomization covers only
the object identity, the in-plane initial position, and the motion
seed; the table, lighting, and camera intrinsics are fixed across all $10{,}000$
episodes so that any cross-environment gap is attributable to the motion
regime alone.

\subsection{Demonstration Generation}
Every episode is collected by an oracle motion-planner solution that has full
access to the simulator state, including the object's true pose and velocity
at every tick. The planner emits a smooth end-effector trajectory at the
control rate of $20\,\text{Hz}$, with each waypoint expressed as a
$6$-D end-effector delta (XYZ translation plus axis-angle rotation) plus a
$1$-D gripper command, exactly the action space exposed to the learned
policy. We discard any episode whose oracle rollout fails the lift criterion
in the first attempt, ensuring that the released dataset contains only
successful demonstrations and that any failure observed during evaluation is
attributable to the policy rather than to an unsolvable initial condition.
The full $1{,}000$-episode pool per environment is provided for downstream
finetuning, but the main benchmark protocol evaluates each method on
$100$ held-out seed indices that lie outside this pool, drawn from a
deterministic seed offset so that all comparisons are over the same trial
realisations.

\subsection{Diagnostic Statistics}
In addition to the per-environment success rate reported in the main text, we
release for each policy run the per-episode trajectory of $(\alpha, K_\text{exec},
v_t, \hat d_t)$ when the policy is wrapped by PPC, the chunk-boundary timestamps
and re-inference cadence, and the minimum gripper--object distance attained
within the $200$-step budget. These finer-grained signals are not used in the
headline numbers but support the analysis in Section~\ref{analysis}; in
particular the empirical $\alpha$ distribution drives our worst-case inference
cost discussion, and the per-object success breakdown verifies that the gain
from PPC is not concentrated on a single target geometry. All statistics will
be released alongside the benchmark code.

\section{Supplementary Experiments and Analysis}\label{supp_exp}

\subsection{Cross-Backbone Comparison of Inference-Time Wrappers} 

In the main results (Table~\ref{tab:baselines}), the comparison wrappers ACT~\citep{act} and BID~\citep{Liu2024BidirectionalDI} are evaluated on the strongest foundational backbone $\pi_{0.5}$ for fairness. We supplement this with a parallel evaluation on a second backbone, SmolVLA~\citep{Shukor2025SmolVLAAV}, to verify that PPC's advantage over chunk-boundary smoothing approaches generalizes across backbones rather than being specific to a single VLA. ACT applies temporal ensembling over overlapping chunks, and BID applies guided rejection sampling at chunk boundaries; both operate without an external dynamics signal. Following the same protocol as Table~\ref{tab:baselines} (100 trials per environment, default deployment configurations), we report per-environment success rate in Table~\ref{tab:appendix_smolvla_wrappers}.

\begin{table}[h]
\centering
\renewcommand{\arraystretch}{0.95}
\setlength{\tabcolsep}{4pt}
\footnotesize
\resizebox{\textwidth}{!}{%
\begin{tabular}{l c ccc ccc ccc cc}
\toprule
\multirow{3}{*}{\textbf{Method}}
& \multirow{3}{*}{\textbf{Static}}
& \multicolumn{3}{c}{\textbf{Uniform Motion}}
& \multicolumn{3}{c}{\textbf{Accelerated Motion}}
& \multicolumn{3}{c}{\textbf{Irregular Motion}}
& \multicolumn{2}{c}{\textbf{Average}} \\
\cmidrule(lr){3-5} \cmidrule(lr){6-8} \cmidrule(lr){9-11} \cmidrule(lr){12-13}
&
& \multicolumn{3}{c}{Moving}
& \multicolumn{3}{c}{Accelerating}
& \multirow{2}{*}{Rand.\,Walk}
& \multirow{2}{*}{Stop\ \&\ Go}
& \multirow{2}{*}{Teleport}
& \multirow{2}{*}{Dyn.\,Only}
& \multirow{2}{*}{All} \\
\cmidrule(lr){3-5} \cmidrule(lr){6-8}
&
& Easy & Med. & Hard
& Easy & Med. & Hard
& & & & & \\
\midrule
SmolVLA (foundational)            & 81 & 76 & 57 & 27 & 41 & 33 & 13 & 53 & 40 & 44 & 42.7 & 46.5 \\
SmolVLA + ACT~\citep{act}         & 73\,\textcolor{red}{\scriptsize($-$8\%)}  & 60\,\textcolor{red}{\scriptsize($-$16\%)} & 49\,\textcolor{red}{\scriptsize($-$8\%)}  & 14\,\textcolor{red}{\scriptsize($-$13\%)} & 52\,\textcolor{green!60!black}{\scriptsize(+11\%)} & 43\,\textcolor{green!60!black}{\scriptsize(+10\%)} & 19\,\textcolor{green!60!black}{\scriptsize(+6\%)}  & 43\,\textcolor{red}{\scriptsize($-$10\%)} & 36\,\textcolor{red}{\scriptsize($-$4\%)}  &  5\,\textcolor{red}{\scriptsize($-$39\%)} & 35.7\,\textcolor{red}{\scriptsize($-$7.0\%)} & 39.4\,\textcolor{red}{\scriptsize($-$7.1\%)} \\
SmolVLA + BID~\citep{Liu2024BidirectionalDI} & 77\,\textcolor{red}{\scriptsize($-$4\%)}  & 73\,\textcolor{red}{\scriptsize($-$3\%)}  & 57\,\textcolor{black}{\scriptsize($\pm 0\%$)} & 18\,\textcolor{red}{\scriptsize($-$9\%)}  & 50\,\textcolor{green!60!black}{\scriptsize(+9\%)}  & 25\,\textcolor{red}{\scriptsize($-$8\%)}  & 18\,\textcolor{green!60!black}{\scriptsize(+5\%)}  & 50\,\textcolor{red}{\scriptsize($-$3\%)}  & 38\,\textcolor{red}{\scriptsize($-$2\%)}  & 48\,\textcolor{green!60!black}{\scriptsize(+4\%)}  & 41.9\,\textcolor{red}{\scriptsize($-$0.8\%)} & 45.4\,\textcolor{red}{\scriptsize($-$1.1\%)} \\
\midrule
\textbf{SmolVLA + PPC (ours)}     & $81^{*}$
& 69\,\textcolor{red}{\scriptsize($-$7\%)}
& 69\,\textcolor{green!60!black}{\scriptsize(+12\%)}
& 58\,\textcolor{green!60!black}{\scriptsize(+31\%)}
& 58\,\textcolor{green!60!black}{\scriptsize(+17\%)}
& 59\,\textcolor{green!60!black}{\scriptsize(+26\%)}
& 35\,\textcolor{green!60!black}{\scriptsize(+22\%)}
& 60\,\textcolor{green!60!black}{\scriptsize(+7\%)}
& 71\,\textcolor{green!60!black}{\scriptsize(+31\%)}
& 53\,\textcolor{green!60!black}{\scriptsize(+9\%)}
& \textbf{59.1}\,\textcolor{green!60!black}{\scriptsize(+16.4\%)}
& \textbf{61.3}\,\textcolor{green!60!black}{\scriptsize(+14.8\%)} \\
\bottomrule
\end{tabular}%
}
\caption{\textbf{Inference-time wrappers on SmolVLA.} ACT and BID degrade dynamic-only success rate ($-7.0$ and $-0.8$ points respectively), while PPC achieves $+16.4$ points. $*$: PPC defaults to the baseline when $v=0$.}
\label{tab:appendix_smolvla_wrappers}
\end{table}

\subsection{Additional Analysis on Experimental Results}

\textbf{Per-backbone gain decomposition.} Among the four foundational VLAs in Table~\ref{tab:baselines}, PPC's largest absolute gain is on GR00T N1.6 ($+28.8$ dynamic-only), followed by $\pi_0$ ($+21.1$), $\pi_{0.5}$ ($+18.6$), and SmolVLA ($+16.4$). This ordering inversely correlates with the backbone's baseline dynamic performance: weaker dynamic baselines leave more room for the closed-form correction to recover. Notably, GR00T N1.6 has the strongest static score ($88\%$) but the weakest dynamic score ($37.3\%$), exhibiting the most severe dynamics blindness and consequently benefiting the most from PPC.

\textbf{ACT's teleport collapse.} ACT on $\pi_{0.5}$ scores $1\%$ on Teleport, a $59$-point drop from the bare backbone ($60\%$). Temporal ensembling maintains a sliding buffer of overlapping chunks and averages them into a single action. When the object teleports, old chunks in the buffer still point toward the pre-teleport position, and averaging actively drags the end-effector away from the new target. The longer the buffer, the longer the stale signal persists. BID avoids this collapse ($48\%$) because rejection sampling discards rather than averages stale chunks. PPC bypasses the issue entirely: the velocity signal detects the teleport via the $\nu_t$ sim-consistency gate, and the wrapper defaults to the unmodified baseline until the next valid reading.

\textbf{DynamicVLA's static regression.} DynamicVLA scores $70\%$ on the static environment, $11$ points below its backbone SmolVLA ($81\%$). Its compact $0.4$B architecture re-infers every $2$ env-steps regardless of whether the scene is changing, injecting inter-chunk discontinuities into an otherwise stable trajectory. Each re-inference resets the action chunk from a slightly different observation, creating micro-jitter that accumulates over a $200$-step episode. This confirms that indiscriminate high-frequency re-planning degrades temporal coherence even when no dynamic compensation is needed, a failure mode that PPC avoids by construction since $\alpha^\star = 1$ and $\delta_k^\star = 0$ when $v = 0$.

\textbf{Irregular motion: regime-specific behavior.} Within the irregular family, Random Walk consistently yields the highest PPC gain across backbones (e.g., $+11\%$ on GR00T, $+7\%$ on SmolVLA), followed by Stop\,\&\,Go, with Teleport last. This ordering reflects the latch stabilizer's operating regime: Random Walk produces frequent direction-shift triggers that keep $L_t$ elevated, enabling the cadence gate to cap $K_\mathrm{exec}$ persistently. Stop\,\&\,Go alternates motion and pause windows, producing intermittent latch activation that partially helps during motion phases. Teleport, by contrast, violates the quasi-stationarity assumption so severely that neither the closed-form correction nor the latch can meaningfully compensate during the discontinuity itself; PPC's gain there comes entirely from improved tracking during the stationary intervals between teleport events.

\textbf{Failure mode: SmolVLA+PPC on Uniform-Easy ($-7\%$).} SmolVLA already reaches $76\%$ on the easiest uniform tier, where object speeds ($1$--$2$\,cm/s) are close to the velocity estimator's noise floor. At these speeds, the cosine-projected $\alpha^\star$ oscillates near $1.0$ with sign noise, occasionally triggering unnecessary compression that shortens the execution window without meaningful tracking benefit. This case represents the lower boundary of PPC's effective operating range rather than a systematic failure.

\section{Supplementary Experimental Details}
\label{sec:appendix_exp_details}

This appendix expands the implementation details summarized in Section~\ref{exp}. All numerical values were extracted directly from the canonical scripts cited per subsection.

\textbf{Hardware.} All foundational VLAs are fine-tuned on a single NVIDIA H200 (141~GB HBM3e). All evaluation rollouts (foundational baselines, comparison wrappers, PPC, and ablations) run on a single NVIDIA RTX~A6000 (48~GB GDDR6). Fine-tuning each backbone takes at most 48 H200-hours; evaluating each method (1{,}000 trials across 10 environments) takes approximately 2 A6000-hours.

\textbf{Foundational VLA Fine-tuning.} Each foundational backbone is fine-tuned from its publicly released checkpoint on the \textsc{MoveBench} demonstration set following the official recipe of the corresponding policy. Table~\ref{tab:appendix_ft_configs} summarizes the per-backbone configuration. All backbones share the same dataset ($\sim$10K demonstrations across the ten environments), the same 7-D action space (6-D arm delta in physical space + 1-D gripper command), and the same control loop ($20$ Hz, \texttt{pd\_ee\_delta\_pose} mode, image resolution and augmentations as in each backbone's released training config). The action chunk length $T_{\rm policy}$ varies by backbone: GR00T~N1.5/N1.6 emits a $16$-step chunk, while $\pi_0$, $\pi_{0.5}$, and DynamicVLA emit $20$-step chunks; SmolVLA and Diffusion Policy use their default chunk size from the released configs. Every backbone's chunked output is consumed end-to-end by the wrapper without modification of the action representation.

\begin{table}[h]
\centering
\renewcommand{\arraystretch}{0.95}
\setlength{\tabcolsep}{6pt}
\footnotesize
\begin{tabular}{@{}l r r r@{}}
\toprule
\textbf{Backbone} & \textbf{Steps} & \textbf{BS} & $\boldsymbol{T_{\rm policy}}$ \\
\midrule
GR00T~N1.5/N1.6  & 20{,}000 & 32 & 16 \\
$\pi_{0}$        & 30{,}000 & 32 & 20 \\
$\pi_{0.5}$      & 30{,}000 & 32 & 20 \\
SmolVLA          & 30{,}000 & 32 & 50 \\
DynamicVLA       & 30{,}000 & 32 & 20 \\
Diffusion Policy & 30{,}000 & 64 & 16 \\
\bottomrule
\end{tabular}
\caption{\textbf{Per-backbone fine-tuning configuration on \textsc{MoveBench}.} BS = batch size; $T_{\rm policy}$ = native chunk length emitted by the policy. All backbones except Diffusion Policy initialize from the official released pretrained checkpoint; optimizer, learning rate, and augmentation schedules follow each backbone's officially released training recipe and are not modified.}
\label{tab:appendix_ft_configs}
\end{table}

\textbf{Comparison Baseline Deployment.} The comparison wrappers ACT and BID are deployed on top of the released $\pi_{0.5}$ checkpoint (and additionally on SmolVLA in Appendix~\ref{supp_exp}) using their original parameters, with one consistency adjustment for fair reactivity: \emph{the execution horizon of every chunk-boundary baseline is capped at 10 env-steps}. Concretely, if a baseline's default execution horizon exceeds 10 it is set to 10, and if its default is below 10 we keep the default. This holds the inter-chunk re-observation latency comparable across methods and prevents any baseline from being penalized by an artificially long open-loop window. The full per-method configurations follow.
\begin{itemize}
    \item \textbf{ACT (temporal ensembling)} on $\pi_{0.5}$: \texttt{decay\_m=0.01}, \texttt{buffer\_size=20}, \texttt{exec\_horizon=1} (ensemble emits one action per env-step, well below the 10-step cap).
    \item \textbf{BID (guided rejection sampling)} on $\pi_{0.5}$: \texttt{n\_samples=4}, \texttt{exec\_horizon=10}, \texttt{overlap\_window=4}.
    \item \textbf{DynamicVLA} uses its own released $0.4$B SmolVLA-based checkpoint fine-tuned on \textsc{MoveBench}. DynamicVLA's defining contribution is a \emph{streaming} pipeline that overlaps action prediction with execution; this streaming protocol is not faithfully reproducible in our SAPIEN simulator (it requires asynchronous prediction running concurrently with simulator time, which our synchronous environment loop does not expose). To preserve DynamicVLA's reactivity argument, we follow its low-latency philosophy at the upper bound permitted by our loop: re-inference every $2$ env-steps, the same minimum cadence as PPC's $K_{\rm exec}$ floor. For reference, running DynamicVLA at the conventional 10-step fallback yields a substantially weaker $31.9\%$ overall ($51/46/44/21/38/37/12/37/28/5$ across Static/Mov.\,E/M/H/Acc.\,E/M/H/RW/SnG/Tele), confirming that the 2-step setting is a strict upper-bound favorable to the baseline.
\end{itemize}

\textbf{PPC Configuration.} All wrapper hyperparameters are fixed across every reported run.
\begin{itemize}
    \item \textbf{Chunk geometry}: $T = 16$, $K = 2$ ($K_{\rm exec}$ floor), $H_{\rm eff} = 10$ ($K_{\rm exec}$ ceiling). For backbones whose native chunk length $T_{\rm policy}$ exceeds 16, only the first 16 model-steps enter the wrapper; remaining model-steps are unused, matching the de-facto execution length set on every comparison baseline ($\le 10$ env-steps consumed per chunk).
    \item \textbf{Latch} (single free knob): $\beta_{\rm in} = 0.3$. The remaining latch constants $\beta_{\rm out}, R_{\rm TH}, L_{\rm th}$ are derived from $(\beta_{\rm in}, K, T)$ per Section~\ref{method}.
    \item \textbf{Bayesian $\alpha$}: $Q = 1.8$, $\beta_{\rm revert} = 0.0$ (random-walk prior, no drift). Under the oracle velocity signal used in simulation, the perception-noise envelope is set to a large sentinel value, forcing $K_{\rm kal} \to 1$ and bypassing Bayesian shrinkage in our experiments. For real-robot deployment, this envelope is supplied by the tracker's confidence stream.
    \item \textbf{Velocity bound}: $V_{\max} = 1$ m/s, matching the xArm6 published maximum TCP velocity. See Appendix~\ref{sec:appendix_supp_ablation} for the robustness check.
    \item \textbf{Grasp gate}: $\|p_{\rm tcp} - p_{\rm obj}\| < 30$ mm (the xArm6 finger half-span); inside this radius the wrapper bypasses, since the object becomes internal state per the contact predicate in Section~\ref{sec:problem}.
    \item \textbf{Workspace}: an $80$~cm world-anchored cube, $[-0.4, 0.4]\!\times\![-0.4, 0.4]\!\times\![0.0, 0.3]$ m.
    \item \textbf{Lift threshold}: $30$ mm above the object's initial height (paper success criterion).
\end{itemize}
Three implementation specifics warrant explicit mention:
\begin{itemize}
    \item \emph{$\Delta p$ source}. The per-step planned delta is computed from the realized TCP trajectory rather than from raw chunk outputs. Specifically, $\Delta p_t = (p^{\rm tcp}_t - p^{\rm tcp}_{t-K})/K$ at every chunk-reset. The first chunk of each episode lacks $K$ prior TCP samples and falls back to $\Delta p_0 = \tfrac{1}{K}\sum_{i=0}^{K-1}\!\mathrm{chunk}[i, :3] \cdot c_{\rm pd}$, where $c_{\rm pd} = 0.04$ is the simulator's controller response factor that maps action-space delta to world-space delta.
    \item \emph{Oracle velocity source}. The dynamics signal $v\hat d$ is obtained from the one-step finite difference of the simulator's ground-truth object position, $v\hat d = \mathrm{cube}_t - \mathrm{cube}_{t-1}$, capped at $V_{\max}$. The simulator's reported velocity field is used \emph{only} as the input to the $\nu_t$ sim-consistency gate (Section~\ref{method}) because that field reads bit-zero on kinematic teleports; the position finite difference is the actual signal driving $\alpha^\star$ and $\delta_k$. On a real robot this signal source would be replaced by a tracker pipeline (e.g., CoTracker3) supplying ($v$, confidence) pairs.
    \item \emph{Channels modified}. The wrapper writes only the $xyz$ translation channel of each chunk action; the $3$-D rotation deltas and $1$-D gripper command are inherited unchanged from the corresponding chunk window. This preserves the policy's grasp-timing decisions and prevents wrapper-induced rotation drift.
\end{itemize}

\textbf{Evaluation Protocol.} For every method (foundational, comparison wrapper, PPC, ablation) we run $100$ trials per environment across the $10$ \textsc{MoveBench} environments, totaling $1{,}000$ trials per method. Trial seeds are $\{0, \ldots, 99\}$ and are disjoint from the demonstration pool. Each trial runs for at most $200$ env-steps ($10$ s of simulator time at $20$ Hz). Success is defined as a successful grasp followed by lifting the object by at least $30$ mm above its initial height (\texttt{FORMAL\_LIFT\_THRESH} $= 0.03$ m). Reported success rates are point estimates; $95\%$ Clopper--Pearson intervals at $n = 100$ correspond to a $\pm 7$\,pp half-width near the $50\%$ region and tighten near the extremes.

\section{Supplementary Ablation Studies}\label{sec:appendix_supp_ablation}

This appendix supplements the closed-form structural ablations and the EMA-stabilizer analysis of Section~\ref{exp} with four additional studies: wrapper computational overhead, adaptive engagement statistics of $\alpha^\star$, the per-env activation rate of the sim-consistency gate $\nu_t$, and a robustness check on the velocity bound $V_{\max}$. All studies use the same protocol as the main ablations (100 rollouts per environment on \textsc{MoveBench}) unless otherwise noted.

\textbf{Wrapper Computational Overhead.} We instrument \texttt{compute\_unified\_correction} on \textsc{MoveBench}-MovingMedium for $5$ episodes ($42$ chunk-resets) and report per-call latency in Table~\ref{tab:appendix_overhead}. The closed-form operator runs in $0.07$\,ms (mean) on a single CPU thread, with $P99$ latency under $0.12$\,ms. Compared to a typical VLA chunk inference of $\sim 64$\,ms for GR00T~N1.6 on an L40 GPU, the wrapper adds $<0.2\%$ overhead, confirming PPC is suitable for real-time deployment without disturbing the inference budget.

\begin{table}[h]
\centering
\renewcommand{\arraystretch}{0.95}
\setlength{\tabcolsep}{6pt}
\footnotesize
\begin{tabular}{@{}l r@{}}
\toprule
\textbf{Statistic} & \textbf{Latency (ms)} \\
\midrule
Mean    & 0.069 \\
Median  & 0.080 \\
P90     & 0.090 \\
P99     & 0.115 \\
Max     & 0.115 \\
\bottomrule
\end{tabular}
\caption{\textbf{Per-chunk PPC computational overhead.} Single-thread CPU latency of \texttt{compute\_unified\_correction} on \textsc{MoveBench}-MovingMedium ($n=42$ chunk-resets).}
\label{tab:appendix_overhead}
\end{table}

\textbf{Adaptive $\boldsymbol{\alpha^\star}$ Engagement.} Section~\ref{method} establishes that $\alpha^\star$ is bounded above by the chunk-budget cap $T/K = 8$ and degenerates to $1$ when no environmental motion is present. Table~\ref{tab:appendix_alpha_dist} reports the per-env distribution of $\alpha^\star$ across all chunk-resets on \textsc{MoveBench}. The distribution is heavy-tailed and regime-adaptive: median $\alpha^\star$ remains near $1$ across all environments while $P90$ scales monotonically with difficulty, reaching the saturation cap of $8$ on AccelHard. Compression engages aggressively only when needed: $\alpha^\star \geq 2$ in $30.2\%$ of AccelHard chunks (vs $5\%$ on Moving) and saturates at $\alpha^\star = 8$ in $10.5\%$ of AccelHard chunks. On Teleporting, $\alpha^\star \equiv 1$ throughout because the $\nu_t$ gate clamps the wrapper to identity on $A_1$-violating chunks.

\begin{table}[h]
\centering
\renewcommand{\arraystretch}{0.95}
\setlength{\tabcolsep}{5pt}
\footnotesize
\begin{tabular}{@{}l r r r r r r@{}}
\toprule
\textbf{Environment} & \textbf{Chunks} & \textbf{Median} & \textbf{P90} & \textbf{Max} & $\boldsymbol{\alpha^\star \geq 2}$ & $\boldsymbol{\alpha^\star = 8}$ \\
\midrule
Teleporting        & 1166 & 1.00 & 1.00 & 1.00 & 0.0\%  & 0.0\%  \\
Moving Easy        & 1014 & 1.00 & 1.64 & 8.00 & 4.7\%  & 0.3\%  \\
Moving Medium      &  987 & 1.01 & 1.60 & 8.00 & 5.2\%  & 0.5\%  \\
Moving Hard        & 1312 & 1.09 & 2.04 & 8.00 & 11.3\% & 1.0\%  \\
Random Walk        & 1846 & 1.00 & 1.74 & 8.00 & 5.5\%  & 0.2\%  \\
Stop and Go        & 1547 & 1.00 & 1.67 & 8.00 & 5.2\%  & 0.3\%  \\
Accelerating Easy  & 1510 & 1.06 & 3.05 & 8.00 & 19.1\% & 3.6\%  \\
Accelerating Med.  & 1822 & 1.11 & 4.86 & 8.00 & 27.4\% & 7.5\%  \\
\textbf{Accelerating Hard} & 2311 & 1.09 & \textbf{8.00} & 8.00 & \textbf{30.2\%} & \textbf{10.5\%} \\
\bottomrule
\end{tabular}
\caption{\textbf{Per-environment distribution of $\alpha^\star$.} Median $\alpha^\star$ stays near $1$ across all regimes; the upper tail scales with difficulty and saturates at $T/K=8$ in $10.5\%$ of AccelHard chunks.}
\label{tab:appendix_alpha_dist}
\end{table}

\textbf{Sim-Consistency Bypass Rate.} The $\nu_t$ gate detects $A_1$ violations from sim-internal contradiction: simulator-reported velocity strictly zero while observed displacement is non-zero. Table~\ref{tab:appendix_nu_t} reports per-environment activation. The gate fires on $82.7\%$ of Teleporting chunks and $0\%$ of all other regimes, exactly matching its design as a regime-specific discrete bypass. The remaining $17.3\%$ of Teleporting chunks correspond to stationary intervals between teleport events where both reported velocity and observed displacement are zero.

\begin{table}[h]
\centering
\renewcommand{\arraystretch}{0.95}
\setlength{\tabcolsep}{6pt}
\footnotesize
\begin{tabular}{@{}l r@{}}
\toprule
\textbf{Environment} & $\boldsymbol{\nu_t = 0}$ \textbf{rate} \\
\midrule
Teleporting              & \textbf{82.7\%} \\
Random Walk              & 0.0\% \\
Stop and Go              & 0.0\% \\
Moving (Easy/Med./Hard)  & 0.0\% \\
Accel. (Easy/Med./Hard)  & 0.0\% \\
\bottomrule
\end{tabular}
\caption{\textbf{Sim-consistency gate $\nu_t$ activation rate.} Active only on Teleporting; dormant on all continuous-dynamics environments.}
\label{tab:appendix_nu_t}
\end{table}

\textbf{$\boldsymbol{V_{\max}}$ Robustness.} The velocity bound $V_{\max} = 1$\,m/s corresponds to xArm6's official maximum TCP velocity. Doubling $V_{\max}$ to $2$\,m/s and re-running AccelHard drops success rate from $33\%$ to $27\%$ (Table~\ref{tab:appendix_vmax}), confirming that exceeding the hardware ceiling produces wrapper aggressiveness that overshoots the physical envelope. We therefore treat $V_{\max}$ as a physical specification rather than a hyperparameter.

\begin{table}[h]
\centering
\renewcommand{\arraystretch}{0.95}
\setlength{\tabcolsep}{6pt}
\footnotesize
\begin{tabular}{@{}l r@{}}
\toprule
$\boldsymbol{V_{\max}}$ & \textbf{AccelHard SR} \\
\midrule
$1$\,m/s (xArm6 spec)   & \textbf{33\%} \\
$2$\,m/s (beyond spec)   & 27\,\textcolor{red}{\scriptsize($-$6\%)} \\
\bottomrule
\end{tabular}
\caption{\textbf{$V_{\max}$ robustness.} Exceeding the hardware specification degrades performance, confirming $V_{\max}$ is a physical constraint.}
\label{tab:appendix_vmax}
\end{table}

\section{Qualitative Visualization}
\label{sec:appendix_qual}

\begin{figure}[h]
    \centering
    \includegraphics[width=\textwidth]{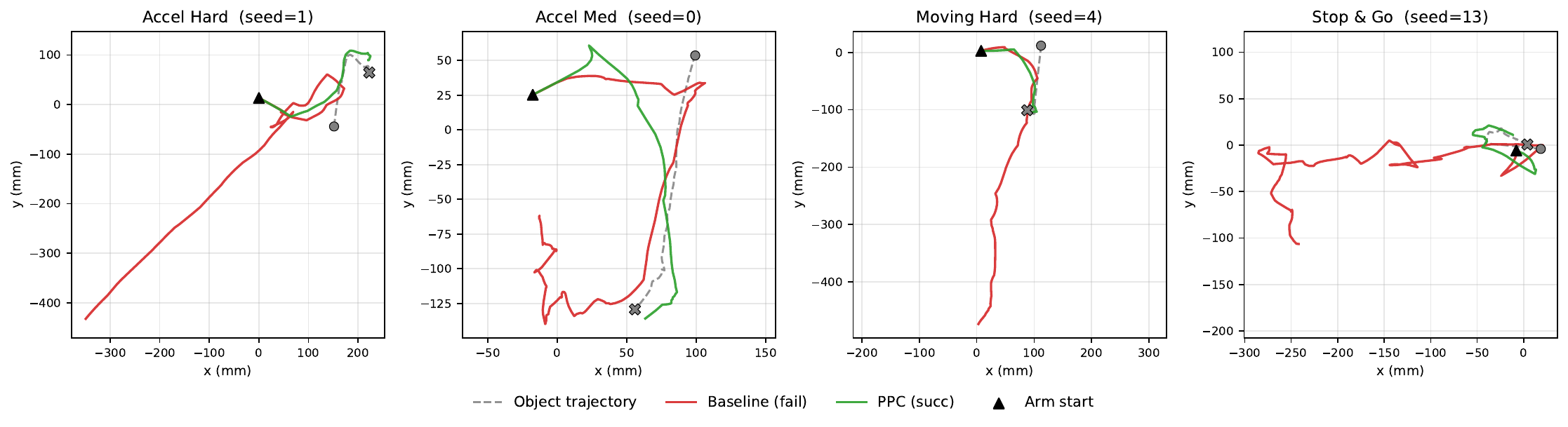}
    \caption{\textbf{Top-down ($x$--$y$) end-effector trajectories on identical seeds.} Gray dashed: object trajectory ($\bullet$ start, $\times$ end). Red: bare baseline TCP (terminates without grasp). Green: PPC-equipped TCP (terminates at grasp). Black triangle: arm start. PPC redirects the chunk-interior path to track the moving target across all four motion regimes.}
    \label{fig:appendix_traj}
\end{figure}

\textbf{Trajectory comparison: baseline VLA vs.\ PPC.} Figure~\ref{fig:appendix_traj} overlays end-effector (TCP) trajectories from the bare baseline and PPC-equipped runs on identical seeds, alongside the corresponding object trajectory, across four representative dynamic environments. In every panel, the baseline TCP veers into the static target's expected position and stops short of the moving object, terminating without grasp at distances of $30$--$100$ mm. The PPC-equipped TCP, sharing the same backbone and seed, instead curves toward the object's current position at the moment of grasp closure, demonstrating that the wrapper actively redirects the executed path within the chunk window without changing the underlying policy.

\begin{figure}[h]
    \centering
    \includegraphics[width=\textwidth]{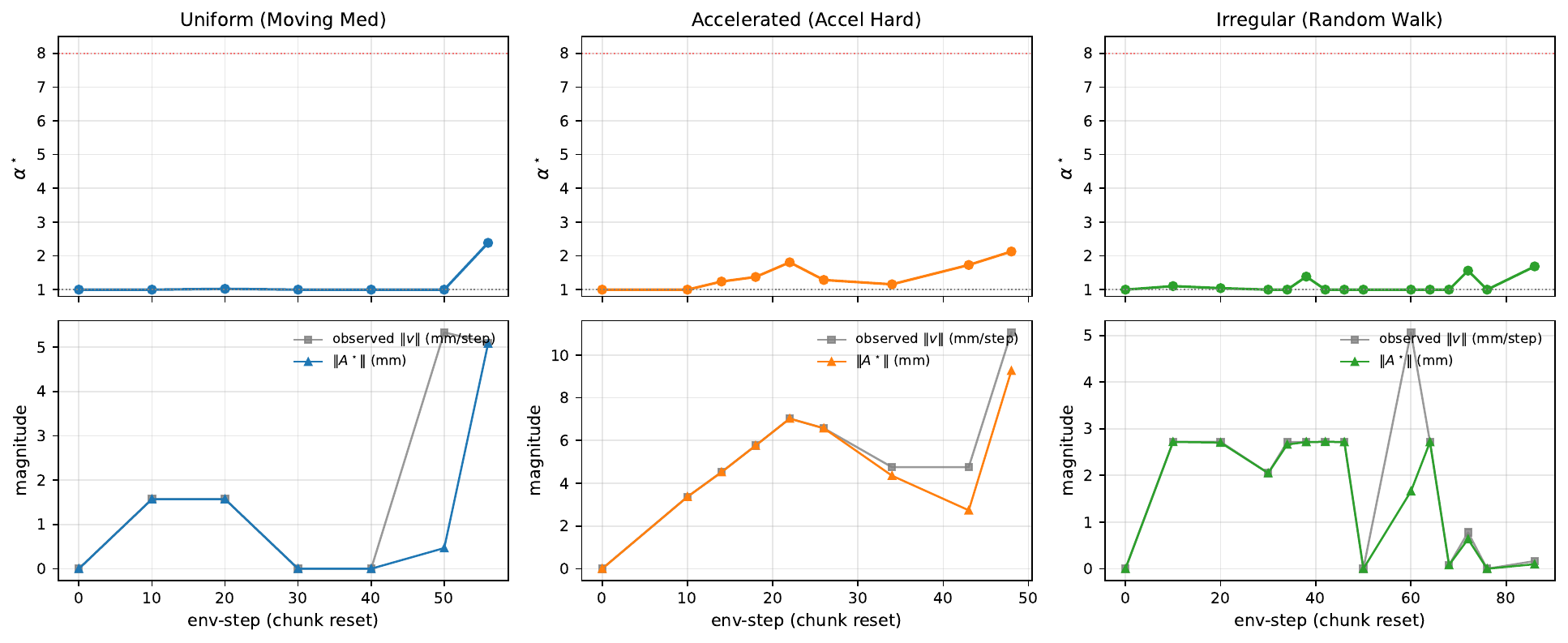}
    \caption{\textbf{Wrapper internals across motion families.} Top row: $\alpha^\star$ per chunk-reset; gray dotted line marks $\alpha=1$ (no compression), red dotted line marks the chunk-budget cap $T/K=8$. Bottom row: observed velocity $\|v\|$ (gray) and disturbance magnitude $\|A^\star\|$ (colored) per chunk. The three regimes produce distinct $\alpha^\star$ profiles: flat near 1 for uniform motion, monotone-rising for accelerated motion, and transient-spiking for irregular motion.}
    \label{fig:appendix_alpha_ts}
\end{figure}

\textbf{Adaptive $\boldsymbol{\alpha^\star}$ engagement across motion families.} Figure~\ref{fig:appendix_alpha_ts} plots per-chunk $\alpha^\star$ and the disturbance magnitudes ($\|v\|$, $\|A^\star\|$) for one successful episode per family (uniform/accelerated/irregular). The three regimes produce qualitatively distinct wrapper signatures. On uniform motion, the observed velocity is small and roughly constant, producing $\alpha^\star\!\approx\!1$ throughout with occasional mild excursions. On accelerated motion, $\alpha^\star$ rises monotonically with the cube's accumulating velocity, saturating at the chunk-budget cap $T/K=8$ in the late chunks where compression matters most. On irregular motion (random walk), $\alpha^\star$ produces transient spikes correlated with direction-shift events, returning to baseline once the regime stabilizes. These traces concretize Section~\ref{method}'s claim that the closed-form $\alpha^\star$ is regime-adaptive: dormant when motion is mild, aggressive when motion is fast, and reactive (rather than chronically engaged) when motion is irregular.

\section{Limitations and Broader Impact}\label{app:limitations}

\subsection{Limitations}

\paragraph{Quasi-stationarity assumption.} The closed-form derivation assumes the disturbance velocity and direction are approximately constant within each executed chunk (A2). While the hierarchical latch mitigates chronic violations by capping $K_{\mathrm{exec}}$, the correction within any single chunk remains suboptimal when the disturbance changes rapidly during execution. Teleportation, the most extreme violation, relocates the object instantaneously and leaves no continuous trajectory for the operator to track, explaining the limited gains observed in that regime.

\paragraph{Simulation-only evaluation.} All experiments are conducted in the ManiSkill simulator with SAPIEN physics. Although the four foundational VLAs tested span diverse architectures, sim-to-real transfer of both the wrapper and the velocity estimation pipeline remains unvalidated. In particular, real-world depth noise, occlusion, and tracker latency may degrade the quality of the external velocity signal beyond the regime where the Bayesian confidence gate can compensate.

\paragraph{Task diversity.} \textsc{MoveBench} isolates motion regime as the evaluation axis through a single task family (pick). While this controlled design is intentional for diagnostic purposes, it leaves open whether PPC's gains transfer to other manipulation primitives such as place, push, or multi-step assembly, where the interaction between chunk-internal dynamics and task semantics may differ.

\paragraph{External velocity signal.} PPC decouples perception from correction by reading velocity from an external source rather than from the VLA backbone. This design sidesteps the ego-motion confound but introduces a dependency on a reliable tracking or depth-sensing pipeline. In cluttered or heavily occluded scenes where object tracking fails, PPC degenerates to the baseline VLA.

\paragraph{Single-object assumption.} The current formulation tracks a single target object per chunk. Multi-object scenes where multiple targets move independently would require either a target-selection mechanism or a multi-channel extension of the cost function, neither of which is addressed in this work.

\subsection{Broader Impact}

This work improves the robustness of robot manipulation policies in dynamic environments without requiring retraining or additional data collection. The primary positive impact is enabling safer and more capable deployment of general-purpose robots in settings where objects or people move during task execution, such as manufacturing lines, household assistance, and human-robot collaboration. Since PPC is a training-free wrapper with no learnable parameters, it does not introduce new data privacy or bias concerns beyond those of the underlying VLA.

On the negative side, more capable manipulation in dynamic settings could lower the barrier for autonomous systems operating in close proximity to humans, where safety-critical failure modes must be carefully addressed before deployment. We emphasize that PPC is evaluated in simulation only and does not constitute a safety-validated system for real-world human-facing applications. Any deployment should include appropriate safety mechanisms independent of the policy layer.



\end{document}